%% file: main.tex
\ifdefined\XeTeXversion\else\pdfoutput=1\fi
\documentclass{article}
\usepackage[T1]{fontenc}
\usepackage{iclr2027_conference,times}
\input{math_commands.tex}

\usepackage{amsmath,amssymb,booktabs,graphicx,microtype,multirow}
\usepackage{xcolor}
\usepackage{hyperref}
\usepackage{url}
\definecolor{linkblue}{RGB}{28,74,140}
\hypersetup{colorlinks=true,linkcolor=linkblue,citecolor=linkblue,urlcolor=linkblue}


\newcommand{\ci}[2]{{\scriptsize\textcolor{black!70}{[#1,\,#2]}}}
\newcommand{\sig}[1]{\textbf{#1}}

\title{When Low Prediction Error\\ Misleads Planning:\\
Diagnosing Representation, Dynamics, and\\ Decision Failures in Latent World Models}

\author{Rui Min$^{2}$, Xianyao Li$^{1}$, Fang Xu$^{1}$ \& Eric Jing Du$^{1,\dagger}$}

\iclrfinalcopy

\begin{document}
\maketitle
\begingroup
\makeatletter
\renewcommand{\thefootnote}{}
\long\def\@makefntext#1{\noindent #1}
\def\Hy@footnote@currentHref{arxiv.authornotes}
\footnotetext{%
$^{1}$X. Li, F. Xu, and E. J. Du are with the Department of Civil and Coastal Engineering, University of Florida, Gainesville, FL 32611, USA.\\
$^{2}$R. Min is with the Department of Mechanical and Aerospace Engineering, University of Florida, Gainesville, FL 32611, USA.\\
$^{\dagger}$Corresponding author: Eric Jing Du, \texttt{eric.du@essie.ufl.edu}.\\
First author: \texttt{rmin@ufl.edu}. Other author e-mails: \texttt{xianyao.li@ufl.edu}, \texttt{xufang@ufl.edu}.
}
\makeatother
\endgroup

\begin{abstract}
The component that dominates a latent world model's prediction error need not be the one whose repair most improves action selection. We show this by comparing action sequences from identical physical starts and separating endpoint error into a candidate-pool center and action-relative responses. Across four model families and four tasks, a confirmation pool of 256 new starts per task and 300 shared candidates per start shows that center error dominates MSE in 14/16 model--task cells. Yet in six of these cells, an oracle that corrects only the action-relative responses yields better physical rank correlation and top-30 elite quality than one that corrects only the center, while leaving more latent MSE (family-wise corrected intervals). The preference differs across the evaluated settings: a separate LeWorldModel (LeWM) study that executes oracle-selected actions favors center repair on PushT and on Reacher with a render-matched goal. Matched-candidate tests localize ordering loss: for LeWM, encoding realized endpoints raises physical Spearman from 0.464 to 0.975 on that Reacher setting and from 0.193 to 0.631 on PushT (64 starts per task), while Cube's encoded-goal cost remains uninformative. A 72-run objective study improves selected response diagnostics, while incremental closed-loop planning gains remain unconfirmed. These results separate error magnitude from the decision effects of oracle correction and motivate evaluating representation, prediction, and planning as separate stages.
\end{abstract}

\begin{figure}[t]
    \centering
    \includegraphics[width=\linewidth,trim=24bp 33bp 27bp 17.5bp,clip]{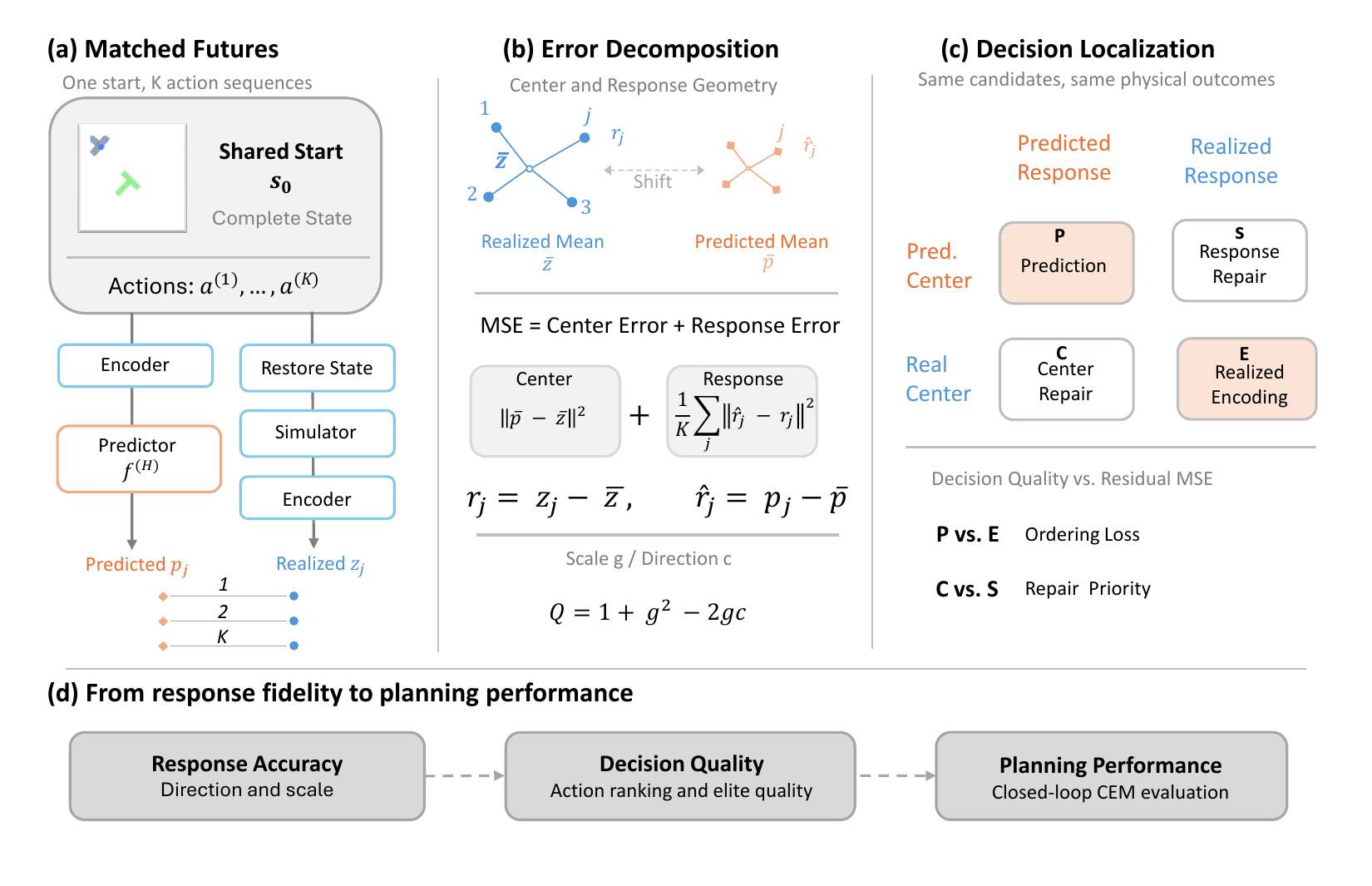}
    \caption{\textbf{Overview of the diagnostic framework.}
    (a) From one restored state, pair predicted and realized outcomes of alternative
    action sequences. (b) Separate center error from action-relative response error,
    then distinguish response scale from direction. (c) On identical candidates,
    compare predicted- versus realized-outcome costs ($P/E$), and center- versus
    response-only oracle repairs ($C/S$). (d) Test transfer from response fidelity
    to physical candidate ranking and elite quality, then closed-loop planning.
    Dashed arrows mark transfers assessed by separate tests.
    Latent geometry is schematic.}
    \label{fig:overview}
\end{figure}

\section{Introduction}

World models encode observations, predict latent consequences of candidate actions, and optimize goal costs~\citep{ha2018worldmodels,hafner2019planet}, spanning joint-embedding predictive architectures~\citep[JEPA;][]{lecun2022jepa,assran2023ijepa,bardes2024vjepa,maes2026lewm}, latent-dynamics planners~\citep{sobal2025pldm}, recurrent models~\citep{hafner2023dreamerv3}, and pretrained visual-feature predictors~\citep{zhou2024dinowm}. Low next-state error measures observed futures, not whether \emph{differences between alternative actions from the same state} or planner cost orderings are correct~\citep{lambert2020objective}.

Observational trajectories can couple state and action: recent motion can reveal the next action, while plausible actions share much of the successor state~\citep{shi2026coco,zhang2026identifiability}. Recent work targets this shortcut~\citep{gan2026actswm,chen2026atm}. A predictor may fit common evolution yet misstate counterfactual direction, scale, or ordering even while using its action input.

Our framework fixes the physical start and varies future actions (Figure~\ref{fig:overview}). Branch centering exactly partitions latent MSE into a center (common-mode) error shared by all candidate actions and an action-relative response error. Complementing training-time decompositions~\citep{zhang2026dwm} and centered readouts~\citep{li2026dueling}, we use this split to diagnose fixed checkpoints. The center term accounts for 62--89\% of LeWM's initial MSE at evaluated horizons; a controlled PushT intervention lowers long-horizon total MSE while \emph{increasing} normalized response error, despite a frozen encoder.

Decision utility adds a distinct test~\citep{wang2026dalewm,chahe2026mcr}. On 256 new starts per task, center error dominates MSE in 14/16 model--task cells across four additional families (80-comparison family-wise intervals). Yet in six cells, an oracle that replaces only the predicted responses with realized ones improves physical ranking and elite quality more than one that replaces only the center, while leaving more MSE (Table~\ref{tab:main-inversion}): the component that dominates the loss need not be the one that matters for decisions. An exploratory study on 64 starts per task compares predicted-outcome cost $P$ with encoded-realized oracle cost $E$ on identical candidates: $E$ improves both decision measures in 9/16 cells. Cube's encoded cost remains uninformative across architectures.

We then train six objectives across four tasks and three paired seeds (72 runs), evaluating rollouts up to H20 and, for three prespecified objectives, closed-loop success with each model's native cross-entropy-method (CEM) planner on 1,024 paired source episodes. Some objectives improve response fidelity, most clearly on Reacher, while incremental cross-task CEM gains over a same-budget base remain unconfirmed. A direction-floor/amplitude-excess split distinguishes response shrinkage from direction recovery, revealing what a lower response loss actually changed (Figure~\ref{fig:shrinkage}).

Our contributions are:
\begin{enumerate}
    \item A matched-candidate test that locates where physical ordering is lost, and a confirmation on 256 new starts per task showing across four model families and four tasks that the error component dominating MSE need not be the one whose correction most improves physical decisions; Cube marks where the encoded-goal metric itself fails.
    \item A same-start action-response fidelity protocol that exposes how aggregate latent MSE can improve while action-relative error worsens; exact center/response and scale--direction decompositions explain why inaccurate directions can reward response shrinkage under squared error.
    \item A controlled 72-run objective study with closed-loop CEM evaluation showing that gains in selected response diagnostics can coexist with statistically unresolved incremental closed-loop gains.
\end{enumerate}

\section{Related work}
\label{sec:related}

Self-supervised learning established effective ways to avoid trivial representations, including BYOL's bootstrapped targets~\citep{grill2020byol} and VICReg's explicit variance--covariance regularization~\citep{bardes2022vicreg}. Decision-aware modeling instead asks which predictive information supports control, as formalized by value-aware losses~\citep{farahmand2017vaml} and value equivalence~\citep{grimm2020valueequiv}. PETS and Dreamer turn learned dynamics into action selection through trajectory sampling and imagined policy training, respectively~\citep{chua2018pets,hafner2023dreamerv3}. These foundations motivate our question: which part of a latent prediction error matters for the decisions made from it?

In latent world models, PhyLatent documents counterfactual local collapse despite LeWM's global regularizer and reports training-time constraints with MPC gains~\citep{maes2026lewm,zeng2026phylatent}. ActSWM, CoCo and ATM target action-dependent prediction~\citep{gan2026actswm,shi2026coco,chen2026atm}. DWM induces a world/action decomposition using an auxiliary world head and orthogonality constraint, while Dueling World Models centers action effects at readout~\citep{zhang2026dwm,li2026dueling}. Our focus is post-training diagnosis: relating these action-relative errors to candidate decisions at fixed checkpoints.

Counterfactual Quotient Models center successor measures; their implemented feature quotient is exact for feature-linear rewards~\citep{chen2026cqm}. For point-latent predictors scored by squared goal distance, a common shift still enters candidate cost gaps. DA-LeWM measures latent--physical Spearman on sampled plans and CEM-stage candidates~\citep{wang2026dalewm}, while MCR directly trains candidate ordering~\citep{chahe2026mcr}. Building on these decision measures, we pair predicted- and observed-outcome costs on identical realized candidates, then compare center-versus-response oracle repair with the MSE budget on new starts.

Especially close, TRM audits an XY rowspace contributing little \emph{terminal latent MSE} on shared TwoRoom candidates~\citep{li2026trm}. We instead repair components of \emph{prediction error} and measure physical ranking and elite quality on the same candidates. The specific contribution is the measured mismatch between MSE dominance and oracle repair priority. Appendix~\ref{app:related-full} details connections to cost repair, output-rule changes, intervention analyses and recurrent-rollout controls.

\section{Problem setup}

Let $o_t$ be an observation, $a_{t:t+H-1}$ an action sequence, $\phi$ an observation-conditioned representation map (encoder or recurrent posterior), and $f$ a latent predictor; the horizon $H$ counts model steps of five environment steps each, and H5 denotes $H=5$. Starting from a shared context $z_0$, the model produces
\begin{equation}
\hat z_H=f^{(H)}(z_0,a_{t:t+H-1}), \qquad z_H=\phi(o_{\leq t+H},a_{<t+H}).
\end{equation}
Standard latent prediction minimizes $\|\hat z_H-z_H\|^2$, usually averaged over observational trajectories. A planner instead samples candidate sequences $a^{(j)}$, evaluates a latent goal cost $d(\hat z_H^{(j)},z_g)$, and selects low-cost candidates~\citep{chua2018pets}. Prediction rewards pointwise agreement on observed futures; planning requires correct \emph{relative} consequences across counterfactual candidates.

\paragraph{Same-start branches.}
For each complete simulator state $s$, we restore the identical state and execute $K$ alternative future action sequences. This yields paired physical trajectories, encoded outcomes $z_j$, and predictions $p_j$. Pairing by complete physical state removes variation in the initial observation, hidden velocity, and contact configuration. Multi-step branches after the first intervention are not described as pure one-step causal effects: their intermediate states legitimately differ because of earlier actions.

\paragraph{Physical criteria.}
Latent agreement is evaluated alongside normalized physical goal discrepancy: object position/angle for PushT, agent position for TwoRoom, joint coordinates for Reacher, and block position for Cube (Appendix~\ref{app:protocol}). These endpoint scores are computed post hoc, not used as latent-goal costs or training targets; because the proxies differ, we do not pool them. In the four-candidate objective study, fixed-pool regret is the mean physical error among exact minimum-latent-cost ties minus the smallest physical error in that start's candidate pool; it is not native CEM success.

\section{Why aggregate prediction error can mislead}
\label{sec:decomp}

\subsection{Exact center/response decomposition}

For one starting state, define branch centers and centered responses
\begin{equation}
\bar z=K^{-1}\textstyle\sum_j z_j,\quad \bar p=K^{-1}\sum_j p_j,
\qquad r_j=z_j-\bar z,\quad \hat r_j=p_j-\bar p.
\end{equation}
Euclidean orthogonality gives
\begin{equation}
\frac1K\sum_j\|p_j-z_j\|^2
=\|\bar p-\bar z\|^2+\frac1K\sum_j\|\hat r_j-r_j\|^2.
\label{eq:decomp}
\end{equation}
This decomposition motivates two same-pool diagnostic oracles: $C_j=p_j-\bar p+\bar z$ replaces only the predicted center, leaving response error; $S_j=z_j-\bar z+\bar p$ replaces only the predicted responses, leaving center error (Section~\ref{sec:repair-priority}). The center denotes the sampled action-pool mean rather than physical null-action dynamics. Under squared Euclidean goal cost, the predicted-versus-$C$ difference in the $i$--$j$ cost gap is $2\langle\bar p-\bar z,p_i-p_j\rangle$ (Appendix~\ref{app:derivation}). The decision effect of center error thus depends on its projection onto candidate differences, rather than its norm alone. An exploratory projection check did not distinguish the six inversion cells of Section~\ref{sec:repair-priority} from other cells (Appendix~\ref{app:repair-priority}); the identity explains how a reversal is possible under this score, while its empirical drivers remain to be established.

\subsection{Response scale and direction}

Stack centered true and predicted responses into $R$ and $\hat R$. When both response energies are nonzero, define
\begin{equation}
g=\frac{\|\hat R\|}{\|R\|},\qquad
c=\frac{\langle\hat R,R\rangle}{\|\hat R\|\|R\|}.
\end{equation}
The normalized response error is
\begin{equation}
Q=\frac{\|\hat R-R\|^2}{\|R\|^2}
=1+g^2-2gc=(1-c^2)+(g-c)^2.
\label{eq:q}
\end{equation}
At fixed direction cosine and nonnegative scale, the least-squares optimum is $g^*=\max(c,0)$. Inaccurate direction can thus reward shrinking the response, a response-level analogue of blurry squared-error video prediction~\citep{mathieu2016multiscale}; forcing $g=1$ can instead expand the wrong response. We retain the two algebraic terms $1-c^2$ and $(g-c)^2$ in our tables. Interpreting the former as a monotone \emph{direction floor} requires $c\geq0$: for $c<0$, the nonnegative-scale minimum is $Q=1$ at $g=0$, not $1-c^2$. Section~\ref{sec:shrinkage} tests amplitude movement toward the least-squares scale rather than equating branch distance with fidelity.

\section{Matched-candidate localization}
\label{sec:matched}

For each start, hold action sequences and physical endpoints fixed; $d$ is the model-native squared Euclidean latent goal score:
\begin{equation}
P_j=d(p_j,z_g),\qquad E_j=d(z_j,z_g),\qquad q_j=\text{physical endpoint error}.
\end{equation}
Here $z_j$ uses candidate $j$'s realized observations: an encoded endpoint for feedforward models and a real-observation recurrent posterior for DreamerV3. We report Spearman with $q_j$, the physical error averaged over selected top-$k$ candidates, and P/E elite overlap; executing the mean action is a separate operator. The four endpoint sets form a center/response $2\times2$ grid: \mbox{$P=(\bar p,\hat r)$, $C=(\bar z,\hat r)$, $S=(\bar p,r)$, $E=(\bar z,r)$}. Weak physical ordering under $E$ points to a bottleneck in the tested representation/goal metric. Stronger ordering under $E$ than $P$ places the loss between prediction and realized-state inference, which includes observation-conditioned posterior updates in recurrent models. Native CEM then tests how these candidate-level differences carry through search and replanning.

\section{Experimental design}

\paragraph{Tasks and model families.}
We study PushT, TwoRoom, Reacher, and Cube. The primary model is LeWorldModel (LeWM), an end-to-end JEPA whose encoder is globally regularized toward an isotropic Gaussian and whose predictor is trained by next-latent regression~\citep{maes2026lewm}. We also test final checkpoints of PLDM~\citep{sobal2025pldm}, PreJEPA (a causal transformer on frozen DINOv2 patch features), DreamerV3~\citep{hafner2023dreamerv3}, and DINO-WM~\citep{zhou2024dinowm} on all four tasks, without new training or checkpoint selection. Rendering, goal and posterior protocols are model specific (Appendix~\ref{app:protocol}), so we compare architectures only through dimensionless within-model quantities.

\paragraph{Candidate pools.}
The C/S confirmation uses 256 new starts per task, shared across the four additional families and disjoint from the exploratory starts, with 300 candidates per start proposed by the final iteration of LeWM's CEM; the comparisons therefore hold the proposer fixed rather than running each model's own CEM (Section~\ref{sec:repair-priority}). The exploratory P/E study reuses 64 physical starts and realized endpoints per task across all five families, again with 300 LeWM-proposed action sequences per start. H5 response geometry uses fixed branch banks of 64 starts per task with $K=4$ branches (Appendix~\ref{app:protocol}).

\paragraph{Objective study.}
Each task's pretrained LeWM encoder and projector are frozen and only latent prediction modules are updated. Six arms share initial weights, branch data, expert batches, optimizer, update budget and three paired seeds: a same-budget squared-error \emph{base}, and objectives adding \emph{direction} calibration, local \emph{branch} geometry, true \emph{transition} increments, an amplitude ablation, or a controlled combination. Each task uses 2,000 branch-training starts and 1,024 evaluation starts with four action candidates plus an independently sampled goal branch. Training uses H5; evaluation measures H1--H20, up to 100 environment steps, over 72 fine-tuning runs. Native CEM evaluates the initial checkpoint and base/transition/branch at three seeds on 256 paired source episodes per task: 10,240 model--episode evaluations but 1,024 distinct source episodes from the original training datasets.

\paragraph{Inference.}
Bootstrap resamples whole paired starts/source episodes rather than individual candidates or branches; fixed-checkpoint intervals condition on those models. The 64 reused starts support exploratory analyses; the fresh starts provide the separate confirmation test. Confirmation uses 100,000 paired-start draws and two-sided Bonferroni percentile intervals with nominal 95\% family-wise coverage over 80 comparisons (Appendix~\ref{app:repair-priority-confirm}). Other comparison families and seeds are in Appendix~\ref{app:protocol}.

\section{Results}

The results form three layers that support different claims. Oracle replacements on fixed candidate pools (Sections~\ref{sec:repair-priority} and~\ref{sec:localization}) show which error component matters for physical action selection and where ordering is lost; the decomposition and response probes (Section~\ref{sec:mechanism}) explain why error magnitude and decision effect can disagree; the objective and closed-loop studies (Section~\ref{sec:shrinkage}) test what learned corrections change. A historical positive control, in which removing LeWM's global regularizer collapses the representation, lowers H5 prediction MSE by four orders of magnitude while closed-loop success falls on every task (Appendix~\ref{app:collapse}); all results below concern healthy checkpoints.

\subsection{MSE dominance does not determine oracle decision sensitivity}
\label{sec:repair-priority}

On 256 new starts per task, 80-comparison family-wise intervals show $\mathrm{MSE}_S>\mathrm{MSE}_C$ in 14/16 cells. One cell has the opposite MSE ordering, and one interval spans zero. Six of the 14 favor $S$ on both physical criteria despite more MSE (Table~\ref{tab:main-inversion}); these cells lie in PreJEPA, DINO-WM and DreamerV3, across PushT, TwoRoom and Reacher. The earlier 64-start sample shows the same qualitative pattern (Appendix~\ref{app:repair-priority}). Table~\ref{tab:sign-grid} shows all 16 cells: no adjusted interval favors $C$ on either decision criterion, and every cell outside Table~\ref{tab:main-inversion} is unresolved on at least one of the three contrasts.

A separate LeWM study that executes the mean action of oracle-selected elites instead favors $C$ on PushT and render-matched Reacher (Appendix~\ref{app:native-operator}), so the preferred correction differs across the evaluated settings. Those two tasks already favor $C$ in fixed-elite point estimates, so executing the elite mean did not create the preference; the two studies differ in models, pools and evaluation operator, and do not isolate which difference matters.

$C$ shares $P$'s branch differences and $S$ shares $E$'s, so $S-C$ adds an MSE axis to the P/E comparison of Section~\ref{sec:localization}; squared-cost center cross-terms can still alter gaps. The added result is the coexistence of better physical decisions and more residual MSE. To separate response shape from energy, SHAPE inserts true centered responses at predicted energy, while AMP rescales predicted responses to true energy. SHAPE wins both criteria in all 12 non-Cube cells under 80-comparison family-wise intervals; Cube's joint result remains unresolved. These oracles are non-deployable diagnostics with unequal true-future information: SHAPE receives the true response vectors, whereas AMP receives their energy.

\begin{table}[!ht]
\caption{\textbf{Six of 16 new-start cells show an MSE--decision inversion.} Four models share 256 starts per task and 300 candidates per start. $\Delta\rho=\rho_S-\rho_C$ and $\Delta q=q_C-q_S$ favor $S$; $\Delta M=\mathrm{MSE}_S-\mathrm{MSE}_C>0$ means more error under $S$. Brackets are nominal 95\% family-wise intervals over 80 comparisons. All 16 cells are summarized in Table~\ref{tab:sign-grid}; full intervals are in Appendix~\ref{app:repair-priority-confirm}.}
\label{tab:main-inversion}
\centering\footnotesize
\setlength{\tabcolsep}{3pt}
\begin{tabular}{@{}llrrr@{}}
\toprule
Model & Task & $\Delta\rho$ & $\Delta q$ & $\Delta M$ \\
\midrule
PreJEPA & PushT & 0.2436 \ci{0.1668}{0.3186} & 0.0982 \ci{0.0386}{0.1662} & 0.3787 \ci{0.3504}{0.4081} \\
PreJEPA & TwoRoom & 0.2626 \ci{0.1816}{0.3437} & 0.0647 \ci{0.0355}{0.0989} & 0.2452 \ci{0.1906}{0.3052} \\
PreJEPA & Reacher & 0.2900 \ci{0.2442}{0.3379} & 0.7475 \ci{0.6279}{0.8742} & 1.0953 \ci{1.0732}{1.1188} \\
DINO-WM & Reacher & 0.2411 \ci{0.1983}{0.2848} & 0.4916 \ci{0.3995}{0.5930} & 1.6051 \ci{1.5822}{1.6288} \\
DreamerV3 & PushT & 0.1852 \ci{0.1264}{0.2428} & 0.1201 \ci{0.0720}{0.1880} & 0.0283 \ci{0.0240}{0.0329} \\
DreamerV3 & TwoRoom & 0.1357 \ci{0.0544}{0.2190} & 0.0410 \ci{0.0061}{0.0825} & 0.0066 \ci{0.0043}{0.0092} \\
\bottomrule
\end{tabular}
\end{table}

\begin{table}[!ht]
\caption{\textbf{All 16 new-start cells.} Signs of the three contrasts of Table~\ref{tab:main-inversion}, in the order $\rho_S-\rho_C$, $q_C-q_S$, $\mathrm{MSE}_S-\mathrm{MSE}_C$. A sign is $+$ or $-$ when the 80-comparison family-wise interval lies wholly above or below zero, and $?$ when it includes zero. In the first two positions $+$ means better decisions under $S$; in the third, $+$ means more residual MSE under $S$, so $+\,+\,+$ is the inversion. Shaded cells are the six inversion cells; no cell has $-$ in a decision position.}
\label{tab:sign-grid}
\centering\footnotesize\setlength{\fboxsep}{1.5pt}
\begin{tabular}{@{}lcccc@{}}
\toprule
Model & PushT & TwoRoom & Reacher & Cube \\
\midrule
PLDM & $?\,+\,+$ & $?\,+\,+$ & $+\,?\,+$ & $?\,?\,+$ \\
PreJEPA & \colorbox{black!12}{$+\,+\,+$} & \colorbox{black!12}{$+\,+\,+$} & \colorbox{black!12}{$+\,+\,+$} & $?\,+\,+$ \\
DINO-WM & $+\,+\,?$ & $?\,?\,+$ & \colorbox{black!12}{$+\,+\,+$} & $?\,?\,+$ \\
DreamerV3 & \colorbox{black!12}{$+\,+\,+$} & \colorbox{black!12}{$+\,+\,+$} & $?\,?\,-$ & $?\,?\,+$ \\
\bottomrule
\end{tabular}
\end{table}

\subsection{Matched candidates localize where physical ordering is lost}
\label{sec:localization}

\begin{figure}[t]
    \centering
    \includegraphics[width=\linewidth]{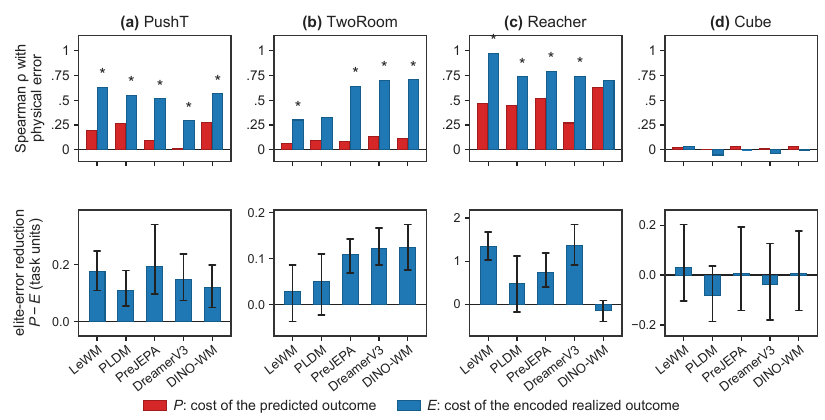}
    \caption{\textbf{Matched candidates expose a task-dependent ordering gap.} $P$ scores the predicted outcome; oracle $E$ scores the model's representation of the realized outcome of the same action sequence. (top) Rank correlation of each cost with physical error on identical candidates. (bottom) Reduction in the physical error of the selected top-30 elites with its family-wise interval; units are task specific, so each task has its own scale. A star marks a rank gain $E-P$ whose family-wise interval excludes zero (eight comparisons for LeWM; separate 32-comparison families for the feedforward and recurrent protocols). PushT recovers on both criteria for all five model families and Cube for none. For LeWM Reacher, the goal image is rendered with the same pipeline as the candidate outcomes (Appendix~\ref{app:protocol}). Exact values are in Appendix Table~\ref{tab:matched}.}
    \label{fig:matched}
\end{figure}

For LeWM, across 64 starts and 300 identical candidates per task, the gap is graded rather than uniform: $E$ improves both ranking and elite quality on PushT and Reacher under eight-comparison family-wise intervals (Figure~\ref{fig:matched}), ranking alone on TwoRoom, and neither on Cube, where both $P$ and $E$ correlations sit near zero. The P/E top-30 overlap ranges from 2.58 to 4.05 candidates, close to the random-overlap expectation of 3, so the two costs often choose nearly disjoint elites.

Figure~\ref{fig:matched} and Appendix Table~\ref{tab:matched} give the contrasts for all five model families. Across PLDM, PreJEPA, DreamerV3, and DINO-WM, $E$ beats $P$ on both criteria in 9/16 cells: all four PushT, three TwoRoom (PreJEPA/DreamerV3/DINO-WM), and two Reacher (PreJEPA/DreamerV3). Together with LeWM, these results distinguish two bottlenecks: ordering present in realized-outcome representations can be degraded by prediction, while Cube's encoded-goal cost remains uninformative across all five families. Its same-pool selection gap persists even under $E$; the retrospective partition in Appendix~\ref{app:protocol} describes that gap without assigning a causal explanation. A descriptive control on the confirmation bank finds 155 Cube starts whose 300 physical scores are all distinct, and the four models' mean $E$ correlations remain between $-0.062$ and $0.064$, so exact physical ties do not explain the Cube result (Appendix~\ref{app:repair-priority-confirm}).

The conditioning controls clarify what P/E localizes. DreamerV3's $E$ uses real intermediate observations and actions, so its gap includes recurrent state inference. Maximum available real history leaves all 12 feedforward P/E classifications unchanged but markedly improves PLDM/Reacher $P$, identifying history availability as a contributor in that arm. Appendix~\ref{app:protocol} reports the control and its separate comparison family.

\subsection{Why the size of center error does not fix its decision effect}
\label{sec:mechanism}

\begin{figure}[t]
    \centering
    \includegraphics[width=\linewidth]{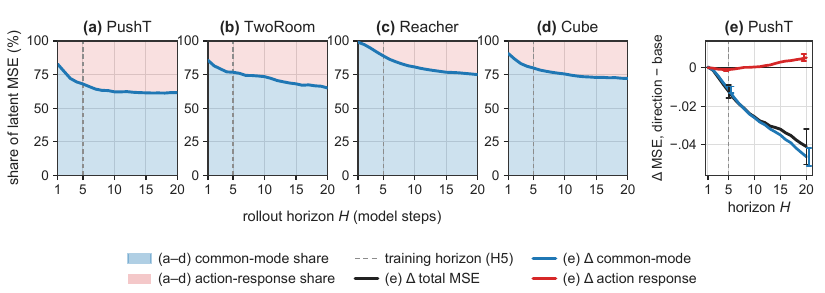}
    \caption{\textbf{Aggregate prediction error can hide worse action response.} (a--d) Share of the initial model's latent prediction MSE carried by the center (common-mode) term against rollout horizon; the remainder is the action-relative response term. (e) PushT, direction objective minus its same-budget base, class means over three seeds: total MSE falls because the center term falls, while the response term rises beyond the training horizon (dashed line). At H5/H20, whiskers are descriptive 95\% intervals for the components and retrospectively frozen 120-comparison family-wise intervals for total MSE.}
    \label{fig:decomp}
\end{figure}

Under squared goal cost, a center shift enters candidate cost gaps only through its projection onto candidate differences (Section~\ref{sec:decomp}), so the size of the center term does not fix its decision effect. Figure~\ref{fig:decomp} reports the decomposition. At H5, the center term accounts for 68.0\% of PushT, 76.7\% of TwoRoom, 88.9\% of Reacher, and 79.9\% of Cube aggregate latent MSE; it remains 61.6--75.0\% at H20. In the PushT direction-versus-base comparison at H20, the change in total state MSE is $-0.041$ \ci{$-0.050$}{$-0.032$}, while the change in $Q$ is $+0.030$ \ci{$0.013$}{$0.049$} under a retrospectively specified 120-comparison family-wise correction (Appendix~\ref{app:protocol}). A $0.046$ reduction in center error masks a $0.0053$ increase in absolute centered-response MSE. This is the only one of the 40 task--contrast--horizon cells with opposite corrected movements in total MSE and $Q$, and it is a response-level mismatch: fixed-pool regret decreases in point estimate with an interval crossing zero, and no cell pairs a corrected MSE gain with a corrected regret loss. Section~\ref{sec:repair-priority} tested the decision-level mismatch directly on matched candidates; Appendix~\ref{app:trajectory} tracks center error and physical ordering along training.

The action probes separate input sensitivity from response fidelity. Across all 20 final model--task checkpoints, mean-action masking or action permutation raises one-step prediction error on 512 fixed probe windows (Appendix Table~\ref{tab:actionuse}); even the weakest, DreamerV3/PushT, worsens by 20\%/34\%. All models therefore use their action input; the branch probes next test whether they use it correctly.

Appendix Table~\ref{tab:geometry} applies the decomposition of Section~\ref{sec:decomp} to same-start branch banks of four further model families. Action-relative response error is 48.9--75.8\% of total H5 latent prediction error and the majority in 15 of 16 arms. Unlike the 300-candidate CEM pools above, these $K=4$ probes sample different actions; error shares are pool-dependent. On TwoRoom, PreJEPA and DINO-WM have near-unit gain (1.007 and 1.048) and \emph{higher} predicted than true branch visibility, yet direction cosines of 0.432 and 0.288 and $Q$ above the zero-response reference $Q=1$. With full recurrent context, DreamerV3 on PushT has gain 1.003 but cosine 0.332; its $Q$ grows from 0.837 at H1 to 1.339 at H5 as the cosine falls from 0.558 to 0.332, consistent with worsening direction agreement along the rollout. PLDM on TwoRoom supplies a healthy positive control ($Q=0.067$, cosine 0.967). Thus visible, correctly scaled branches can still point in the wrong latent direction: accurate action geometry requires direction as well as separation and gain.

\subsection{Objectives: response shrinkage and the closed-loop boundary}
\label{sec:shrinkage}

Squared error can lower $Q$ by shrinking responses rather than by correcting their direction. On PushT at H20, squared-error fine-tuning lowers the gain from 0.71 to 0.25: the base-minus-initial amplitude-excess contrast is $-$0.274 \ci{$-$0.332}{$-$0.222}, whereas the direction-floor contrast of $+$0.007 \ci{$-$0.004}{0.019} spans zero. Other tasks do improve direction, and direction-plus-log-amplitude supervision improves PushT's floor, although its fixed-pool regret interval spans zero (Appendix~\ref{app:shrinkage}).

\paragraph{Preregistered response--decision comparison.}
A preregistered replication compares the squared-error and non-shrinkage objective classes on 2,048 fresh PushT starts, with four predictions fixed before these starts were collected. It reproduced the predicted H20 response ordering, with $Q$ favoring squared-error objectives, but the fixed-pool regret difference was not resolved and the H5 $Q$ prediction was contradicted; the sample size was set after the exploratory results but before the new data (Appendix~\ref{app:stage-b}). Changing the training objective thus altered response geometry without a confirmed change in fixed-pool decisions.

\begin{figure}[!ht]
    \centering
    \includegraphics[width=\linewidth]{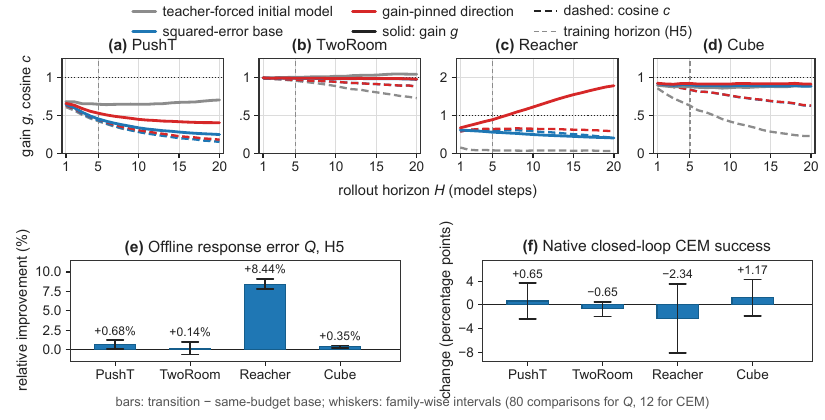}
    \caption{\textbf{Response shrinkage and the closed-loop conversion boundary.} (a--d) Gain $g$ (solid) and direction cosine $c$ (dashed) versus rollout horizon: one initial checkpoint and trained-arm means over three paired seeds on 1{,}024 held-out starts. The vertical line marks training H5; the initial Reacher gain ($\approx$3.8, an observation-domain shift) is above the plotted range. The identity $Q=(1-c^2)+(g-c)^2$ separates a direction floor (for $c\geq0$) from amplitude excess. Squared-error fine-tuning moves gain toward cosine; pinning gain over-expands on Reacher beyond H5. (e,f) Transition minus same-budget base: H5 $Q$ improvement and native CEM success change, with family-wise intervals (80 and 12 comparisons). Every CEM interval includes zero; task scores are not pooled.}
    \label{fig:shrinkage}
\end{figure}

\paragraph{Closed-loop planning.}
Reacher transition improves offline H5 response error by 8.44\% but changes CEM success by $-2.34$ points [${-8.07},3.52$] (Figure~\ref{fig:shrinkage}e,f). All corrected transition-/branch-versus-base native-CEM success intervals cross zero. Separately, PushT base is $-6.64$ points below initial [${-11.98},{-1.43}$] (Appendix Table~\ref{tab:cem}); that contrast measures the effect of shared fine-tuning rather than an added response loss. A repeat-to-varying action-block control lowers offline H5 $Q$ on PushT, Reacher and Cube, while its eight corrected native-CEM success intervals cross zero (Appendix~\ref{app:protocol}).

Recursive use of predictions adds a further source of sensitivity~\citep{talvitie2017selfcorrecting}; our equal-norm continuation probes and their negative controls (Appendices~\ref{app:continuation}--\ref{app:direction-specificity}) reproduce the same signature with real-trajectory directions, so it is not specific to predictor error.

\section{Limitations and conclusion}

The largest component of latent MSE can be the less valuable component to repair for action selection. Our new-start confirmation demonstrates this mismatch on identical candidate pools: response repair can improve physical ranking and elite quality more than center repair while leaving more prediction error. The scale--direction analysis explains why squared-error objectives can favor smaller responses even when their direction remains inaccurate. Together, these findings give world-model evaluation a concrete structure: test which physical distinctions the representation retains, which survive prediction, and which the planner converts into better choices.

The preferred correction varies across the evaluated settings; the true-future oracles are diagnostic tools rather than deployable corrections. The confirmation extends start coverage within the studied tasks, while using a common proposer leaves generalization to other candidate distributions for future work. The objective-study and native-CEM evaluations draw on existing datasets, and possible pretraining overlap further limits claims about unseen trajectories. The objective studies make the next challenge concrete: learn response corrections from available observations and actions, without oracle access at decision time, and test their incremental benefit under matched-budget closed-loop planning.
\clearpage
\subsection*{AI use statement}
The research questions, diagnostic framework, analyses and conclusions in this work are the authors' own. We used generative AI tools, under the authors' direction, as execution and writing aids: to support research execution, to retrieve and identify related work, to draft sections of the manuscript, and to polish the writing. We did not use generative AI tools to generate synthetic datasets, to clean or reformat datasets, or to formulate or prove mathematical claims; translation and qualitative or thematic data analysis are not applicable to this work. We have reviewed all AI-assisted work: references were checked against primary sources and reported results were checked against recorded experimental outputs. We take responsibility for the final content of this work, including all AI-assisted text, claims, and artifacts.

\subsection*{Ethics statement}
This work uses simulated control environments and existing research models. It contains no human-subject data and introduces no deployment involving people. We report negative and unresolved results to reduce selective-reporting risk and distinguish privileged diagnostic signals from deployable methods.

\subsection*{Reproducibility statement}
Appendix~\ref{app:protocol} specifies experimental units, pairing, horizons,
comparison families and resampling settings; Appendices~\ref{app:repair-studies}
and~\ref{app:stage-b} give the full protocols of the oracle repair studies and
the preregistered replication. The supplementary bundle contains the
manuscript source, analysis code, frozen study protocols and saved per-start
statistics for the new-start confirmation and the exploratory repair study,
with CPU scripts that recompute their family-wise contrasts and the LeWM P/E
means. It does not include raw images, model weights, training runs, simulator
replays or full native-CEM histories; its README states which results can be
recomputed from the bundle and which can be traced to audited summaries.

\clearpage
\bibliography{references}
\bibliographystyle{iclr2027_conference}

\clearpage
\appendix

\input{related_work_appendix}

\input{derivation_appendix}

\input{protocol_appendix}

\input{training_trajectory_appendix}

\input{extended_tables_appendix}

\clearpage
\input{response_priority_appendix}

\input{native_operator_appendix}

\clearpage
\input{objective_appendix}

\input{stage_b_appendix}

\clearpage
\input{continuation_appendix}

\input{continuation_secant_appendix}

\input{residual_origin_appendix}

\clearpage
\input{collapse_appendix}

\end{document}

%% file: math_commands.tex
\usepackage{amsmath,amsfonts,bm}

\def\eqref#1{equation~\ref{#1}}

\def\1{\bm{1}}

\DeclareMathAlphabet{\mathsfit}{\encodingdefault}{\sfdefault}{m}{sl}
\SetMathAlphabet{\mathsfit}{bold}{\encodingdefault}{\sfdefault}{bx}{n}



%% file: related_work_appendix.tex
\section{Extended related-work distinctions}
\label{app:related-full}

Our decomposition presupposes a representation worth predicting, continuing collapse-avoidance work in self-supervised learning~\citep{grill2020byol,chen2021simsiam,zbontar2021barlow,bardes2022vicreg}. LeWM regularizes embeddings toward a global isotropic Gaussian~\citep{maes2026lewm}, while LpWM studies sparse, non-Gaussian codes~\citep{kuang2026lpwm}. More directly, PhyLatent documents physical and counterfactual local collapse despite LeWM's global regularizer and reports training-time constraints with MPC gains on the same four task names~\citep{zeng2026phylatent}. We therefore take local non-collapse and branch separation as given and ask which component of a fixed checkpoint's error matters for physical candidate decisions.

A second line targets this counterfactual structure directly: ActSWM supervises predicted transitions to be separated and action-recoverable over long horizons~\citep{gan2026actswm}; CoCo imposes counterfactual consistency across inverse-action, zero-action, and mirrored rollouts to suppress visual-inertia shortcuts~\citep{shi2026coco}; and ATM compares action information in encoded against predicted transitions~\citep{chen2026atm}. Branching web-agent data and predicted-state matching have also been used to distinguish alternative-action outcomes and improve action ranking~\citep{li2026webdwm}. Closer to our split, DWM adds an auxiliary world head with an action-invariance objective and couples it to the original prediction head through an orthogonality constraint, inducing complementary world and action components~\citep{zhang2026dwm}. Dueling World Models subtracts each prediction's mean action effect at readout; its gridworld control gain does not carry to its DMC CEM test~\citep{li2026dueling}. Our use of the split is diagnostic rather than architectural: we decompose a fixed aggregate loss after training and test whether a controlled intervention lowers its common-mode term while raising its action-relative term (Section~\ref{sec:decomp}).

Counterfactual Quotient Models center successor measures, preserving comparisons for bounded rewards in theory; their implemented feature quotient is exact for feature-linear rewards~\citep{chen2026cqm}. We instead diagnose existing point-latent predictors scored by squared goal distance: a common latent shift is not an additive cost baseline, because it enters candidate cost gaps through a center--response cross term.

A third line sidesteps fidelity and re-aligns the planning cost with decisions directly, as value-aware and value-equivalent models do~\citep{farahmand2017vaml,grimm2020valueequiv}: DA-LeWM measures latent--physical rank correlation on sampled and CEM-stage candidates~\citep{wang2026dalewm}, Monotone Cost Ranking supervises candidate ordering for image-goal navigation~\citep{chahe2026mcr}, TD-JEPA mines a temporal-progress cost from demonstration ordering~\citep{bai2026tdjepa}, and value-guided JEPA reshapes latent distance toward a goal-conditioned value~\citep{destrade2026value}. TwoRoom cost repairs on frozen LeWM, including TRM's shared-candidate audit of an XY rowspace contributing little \emph{terminal latent MSE}, precede our \emph{prediction-error} repair test~\citep{singh2026objective,li2026trm}; LEAP instead adds a learned terminal readout, action proposal and refinement to the released LeWM~\citep{pham2026leap}. ASAR changes the output rule instead: action-neighborhood reconstruction improves fixed-LeWM Cube execution, but its unretuned Reacher/PushT transfers are unresolved~\citep{li2026asar}. These gains show why our fixed-checkpoint oracle comparisons are not impossibility results for downstream planning. Our matched-candidate oracle (Section~\ref{sec:matched}) diagnoses an \emph{existing} predictor and native latent cost by testing whether the observed-outcome cost retains physical ordering that the open-loop cost loses on the same actions and outcomes.

Decision-centric ladders separate prediction from utility~\citep{yu2026decisioncentric}; Dual-Frontier proves passive non-identifiability of fixed-operator agent/world-model return attribution~\citep{zhu2026dualfrontier}. Our narrower test asks which latent error component an oracle should repair for physical ranking and elite choice on the same candidates, without return guarantees.

The Intervention Gap already separates task-query capture from matched effect propagation in LeWM and PreJEPA~\citep{vakalis2026intervention}; our P/E and C/S tests instead compare physical rank and elite outcomes on the \emph{same} candidates.

Self-correction, Jacobian regularization, attenuation with information retention, and value-gradient weighting are established~\citep{talvitie2017selfcorrecting,fang2025generalization,li2026koopmandreamer,voelcker2023vagram}. Recent work studies planner-reachable prediction error, latent transport, and recursive error in SG-JEPA~\citep{you2026control,zhang2026propagation,liu2026semigroup}. Our equal-norm continuation probes (Appendix~\ref{app:mechanism}) build on this work with a narrower question about the direction of the error.

%% file: derivation_appendix.tex
\section{Derivation of the decompositions}
\label{app:derivation}

Let $e_j=p_j-z_j$ and $\bar e=K^{-1}\sum_j e_j=\bar p-\bar z$. Then
\begin{align}
K^{-1}\sum_j\|e_j\|^2
&=K^{-1}\sum_j\|\bar e+(e_j-\bar e)\|^2\\
&=\|\bar e\|^2+K^{-1}\sum_j\|e_j-\bar e\|^2,
\end{align}
because the cross term is proportional to $\sum_j(e_j-\bar e)=0$. Since $e_j-\bar e=\hat r_j-r_j$, this yields Equation~\ref{eq:decomp}. Equation~\ref{eq:q} follows by expanding the squared norm and substituting the scale and cosine definitions. Completing the square gives $Q=(1-c^2)+(g-c)^2$; minimizing over $g\geq0$ gives $g^*=\max(c,0)$.

For the squared Euclidean goal score used in the fixed-pool oracle tests, write $d=\bar p-\bar z$ and $C_j=p_j-d$. If predicted response energy vanishes while true response energy is nonzero, then $g=0$ and $Q=1$ but $c$ is undefined, and no direction is imputed. The difference between the predicted and center-repaired cost gaps of candidates $i$ and $j$ is exactly
\begin{equation}
\bigl(\|p_i-z_g\|^2-\|p_j-z_g\|^2\bigr)
-\bigl(\|C_i-z_g\|^2-\|C_j-z_g\|^2\bigr)
=2\langle d,p_i-p_j\rangle .
\end{equation}
The common $-\|d\|^2$ and goal terms cancel between candidates. If $d$ is orthogonal to the span of all predicted candidate differences, center repair changes their absolute costs but none of their pairwise orderings, regardless of $\|d\|^2$; if it is aligned, center repair can change them. The identity separates the size of a center error from its effect on candidate ordering under squared Euclidean goal cost. Which component matters empirically is then tested by the matched-candidate repairs in Appendix~\ref{app:repair-studies}.

%% file: protocol_appendix.tex
\section{Experimental protocols and statistical design}
\label{app:protocol}

\paragraph{Reproduction materials.}
The supplementary source package contains protocols, code and saved statistics; its README specifies the included data, replay commands and external-data requirements.

\paragraph{Objective study.}
Each task has 2,000 branch-training starts, $K=4$, and training horizon H5. Evaluation has 1,024 starts, four candidates, a fifth independent goal branch, and H1--H20 readouts. One model step spans five environment steps. The six objectives and three paired seeds give 18 trained checkpoints per task. The primary comparison family contains 80 prespecified H5/H20 response and regret contrasts. The separate 120-comparison prediction--decision analysis was frozen before its recomputation but after some H20 outcomes were known; it is retrospective, not an independent confirmation.

\paragraph{Physical endpoint scores.}
With $\Delta$ denoting endpoint minus goal coordinates, the physical error proxies are
\[
\begin{aligned}
q_{\mathrm{PushT}}&=\max\!\left\{\frac{\|\Delta b_{xy}\|_2}{20},
  \frac{|\operatorname{wrap}_{[-\pi,\pi)}(\Delta b_\theta)|}{\pi/9}\right\},\\
q_{\mathrm{TwoRoom}}&=\frac{\|\Delta x_{xy}\|_2}{16},\qquad
q_{\mathrm{Reacher}}=\frac{\|\Delta q_{\mathrm{joint},1:2}\|_\infty}{0.05},\\
q_{\mathrm{Cube}}&=\frac{\|\Delta b_{xyz}\|_2}{0.04}.
\end{aligned}
\]
Here $b_{xy},b_\theta$ are PushT object pose, $x_{xy}$ is TwoRoom agent
position, $q_{\mathrm{joint},1:2}$ are Reacher joint coordinates, and
$b_{xyz}$ is Cube block position. The normalized proxies are dimensionless
but have different task meanings. Fixed-pool top-30 elite error averages
the proxy over selected individual candidates; it is not the physical result
of executing their mean action sequence.

\paragraph{Matched candidates.}
Each task uses 64 starts and 300 candidate action sequences, with top-30 elites. P and E always score the same candidate and are compared with the same physical endpoint. Eight simultaneous contrasts cover rank gain and elite-error reduction on four tasks. For LeWM Reacher, the goal image is rendered with the same pipeline as the candidate outcomes; using the dataset goal image instead leaves the P/E classification unchanged.

\paragraph{Action-use control.}
For each model--task pair we fix 512 probe windows (Table~\ref{tab:resampling-settings}) and evaluate the final checkpoint only. The windows come from each model's held-out validation split, except for DreamerV3, which was trained on all converted episodes without a held-out set. Masking replaces the action by the dataset-mean action, which is zero in the normalized action space of the JEPA-style models and the raw mean for DreamerV3, whose adapter consumes unnormalized actions; shuffling permutes whole action windows across the probe set with a fixed seed. PreJEPA and DINO-WM replace the action at all three context positions and score the pixel-embedding sub-block only, as in their training losses; DreamerV3 imagines one open-loop macro-step from the true posterior at each context position. Each cell of Table~\ref{tab:actionuse} is the loss increase divided by the true-action loss. The control is descriptive: one checkpoint per arm, no resampling interval, and magnitudes are not comparable across architectures because each loss lives in a different latent space. DreamerV3 contributes its single exported final checkpoint.

\paragraph{Cross-family geometry and matched candidates.}
Geometry uses fixed branch banks of 64 starts, $K=4$, and H5, with 10,000 start-level bootstrap resamples; its intervals are descriptive and not family corrected. PLDM sees full visual context and DreamerV3 full recurrent context. The matched-candidate contrast reconstructs each complete 25-step action plan from the saved native-CEM candidates with the original action scaler, because many candidates terminate early, and requires the executed prefix to agree with the audited physical trajectory. The feedforward contrast uses 100,000 paired start-level resamples and two-sided family-wise 95\% intervals over 32 contrasts. A separate 24-comparison family tests the direct effect of maximum available real history; 61/64 PushT, 49/64 TwoRoom, 59/64 Reacher, and 61/64 Cube starts contain all three context frames, and shorter histories use only genuine frames. All 12 feedforward P/E classifications are unchanged, while PLDM/Reacher $P$ improves from $\rho=.444$ to $.718$ and elite error $2.066$ to $1.142$. Writing $q_O$ for the mean physical error of the physically best 30 candidates in the same pool, the elite gap splits exactly as $q_P-q_O=(q_P-q_E)+(q_E-q_O)$. In a retrospective analysis of Cube, $q_E-q_O$ is .316--.433 across the five families while $|q_P-q_E|\leq.082$: even encoded realized outcomes leave most of the selection gap. These signed terms are descriptive, not causal shares. The Dreamer contrast replays 76,800 candidate trajectories, updates its posterior at every real macro observation, and uses a separate prespecified 32-comparison family. No model is trained, no candidate is regenerated, and no CEM is run.

\paragraph{Native CEM.}
Each task uses 256 independent source episodes paired across the initial checkpoint and three seeds each of base, transition, and branch. The 12 planned contrasts comprise transition--base, branch--base, and base--initial for four tasks. Bootstrap sampling units are source episodes. Model--episode rows sharing a source episode are not treated as independent.

\paragraph{Auxiliary action-block coverage control.}
A separately specified $2\times2$ data/objective intervention trains repeated and within-block-varying action sequences with either base or transition loss on each task (three paired seeds, 500 updates each). Both action families share 2,000 physical starts per task, prefixes and first micro-actions. Varying the later micro-actions also changes the resulting physical responses, so this is a data intervention rather than an isolated change in direction error. Evaluation uses the prespecified varying-action distribution at H5 and 256 paired native-CEM source episodes per task. The offline and closed-loop results, with their 80- and 24-comparison corrections, are reported in Table~\ref{tab:coverage-control}.

\paragraph{Randomization and interval construction.}
Table~\ref{tab:resampling-settings} collects the numerical resampling settings used by the oracle comparisons and mechanism probes. Each study retains its own prespecified comparison family; later exploratory tests are interpreted in their chronology rather than combined into a retrospective global family.

\input{protocol_settings}

%% file: protocol_settings.tex
\begin{table}[!htbp]
\caption{Resampling settings for the preregistered, oracle and exploratory studies. Each row is a separate comparison family with nominal 95\% family-wise coverage; the two-sided percentile interval uses tail probability $0.05/(2m)$ for $m$ comparisons. Starts or episodes are paired before resampling; a group contains four starts linked by donor matching.}
\label{tab:resampling-settings}
\centering\footnotesize
\setlength{\tabcolsep}{4pt}
\begin{tabular}{@{}p{5.0cm}lrrl@{}}
\toprule
Study & Unit & Draws & $m$ & Bootstrap seed \\
\midrule
Preregistered replication (Appendix~\ref{app:stage-b}) & Start & 20,000 & 4 & 22092041 \\
Exploratory repair (Appendix~\ref{app:repair-priority}) & Start & 100,000 & 96 & $744021+i_c$ \\
New-start repair (Appendix~\ref{app:repair-priority-confirm}) & Start & 100,000 & 80 & $748321+i_t$ \\
\midrule
Directional continuation (Appendix~\ref{app:continuation}) & Start & 50,000 & 24 & 730021 \\
Condition-matched directions & Group & 50,000 & 32 & 735021 \\
True-history control & Group & 50,000 & 16 & 736021 \\
Restricted Flow-JEPA extension & Window & 50,000 & 16 & 731021 \\
Fixed-reference objective & Start & 50,000 & 32 & $737021+i_t$ \\
Fixed-reference native CEM & Episode & 50,000 & 8 & $738071+i_t$ \\
\midrule
Secant projection (Appendix~\ref{app:secant}) & Group & 50,000 & 8 & 739021 \\
Matched component interventions & Group & 50,000 & 32 & 740021 \\
Energy-matched random decompositions & Group & 50,000 & 16 & 741021 \\
True-history residuals (Appendix~\ref{app:direction-specificity}) & Group & 50,000 & 16 & 742021 \\
Persistence and centroid controls & Group & 50,000 & 32 & 743021 \\
\bottomrule
\end{tabular}

\smallskip
\begin{minipage}{\linewidth}\footnotesize
Task index $i_t=0,1,2,3$ denotes PushT, TwoRoom, Reacher and Cube.
Exploratory-study case index $i_c$ follows the fixed model--task order in the released protocol.
Other fixed randomization seeds are 42 for action-use windows, 735001 for donor grouping,
and 741031--741033 for the three averaged Gaussian decomposition draws.
These numerical seeds are collected here to keep the experiment narratives focused on their design and results.
\end{minipage}
\end{table}

%% file: training_trajectory_appendix.tex
\section{Prediction error and candidate ranking along training}
\label{app:trajectory}

We examine how prediction error and candidate ranking change across the ten saved epochs of each LeWM run. From epoch 1 to epoch 10, normalized prediction error falls by 82\% on PushT, 68\% on TwoRoom and 28\% on Cube. Its center (common-mode) term falls by 88\%, 75\% and 32\%, while response error $Q$ falls by 22\%, 30\% and 1\%, respectively. These different rates motivate tracking physical ordering alongside the loss decomposition.

This training-free diagnostic asks whether falling prediction error buys the ordering a planner consumes. For epochs 1--10 of the four LeWM runs that the main study starts from, we recompute, with each epoch's own encoder and predictor, the matched-candidate quantities of Section~\ref{sec:matched} on the frozen 64 starts and 300 candidates, together with the decomposition of Section~\ref{sec:decomp} over the same candidates. Errors are ratios of sums over starts, normalized by the true within-start response energy, so that total error equals its center term plus $Q$. The plan, including the five contrasts between epoch 10 and epoch 1 and their 20-comparison family, was frozen before execution. As an implementation check fixed in advance, the epoch-10 checkpoint had to reproduce the published LeWM rank correlations within .02: it does for PushT, TwoRoom, and Cube (largest deviation .005), and its predicted endpoints agree with the native planner's saved predictions to a median relative error below .2\%. Reacher reproduces $\rho_P$ but not $\rho_E$, because this pipeline encodes the natively rendered endpoints rather than with the render-matched goal used for the LeWM Reacher row; its encoded-outcome and error curves are therefore shown for completeness only.

\begin{table}[!htbp]
\caption{Change from epoch 1 to epoch 10 on the frozen candidates, with 20-comparison family-wise intervals. Bold marks an interval that excludes zero. $\dagger$: protocol differs from the LeWM Reacher row (see text).}
\label{tab:trajectory}
\centering
\small
\setlength{\tabcolsep}{4pt}
\begin{tabular}{@{}lcccccccc@{}}
\toprule
 & \multicolumn{2}{c}{\textbf{$\Delta\log$ total error}} & \multicolumn{2}{c}{\textbf{$\Delta Q$}} & \multicolumn{2}{c}{\textbf{$\Delta\rho_P$}} & \multicolumn{2}{c}{\textbf{$\Delta\rho_E$}} \\
\cmidrule(lr){2-3}\cmidrule(lr){4-5}\cmidrule(lr){6-7}\cmidrule(l){8-9}
Task & est. & CI & est. & CI & est. & CI & est. & CI \\
\midrule
PushT & \sig{$-$1.719} & \ci{$-$2.209}{$-$1.283} & \sig{$-$.253} & \ci{$-$.524}{$-$.043} & \sig{.132} & \ci{.000}{.262} & .036 & \ci{$-$.036}{.109} \\
TwoRoom & \sig{$-$1.149} & \ci{$-$1.321}{$-$.974} & \sig{$-$.888} & \ci{$-$1.579}{$-$.413} & .046 & \ci{$-$.067}{.158} & $-$.147 & \ci{$-$.386}{.074} \\
Reacher$^\dagger$ & \sig{.192} & \ci{.082}{.314} & \sig{17.773} & \ci{15.718}{20.068} & \sig{.466} & \ci{.299}{.627} & \sig{.293} & \ci{.087}{.487} \\
Cube & \sig{$-$.330} & \ci{$-$.575}{$-$.122} & $-$.017 & \ci{$-$.142}{.104} & .022 & \ci{$-$.017}{.070} & \sig{.093} & \ci{.011}{.189} \\
\bottomrule
\end{tabular}
\end{table}

\begin{figure}[!htbp]
    \centering
    \includegraphics[width=\linewidth]{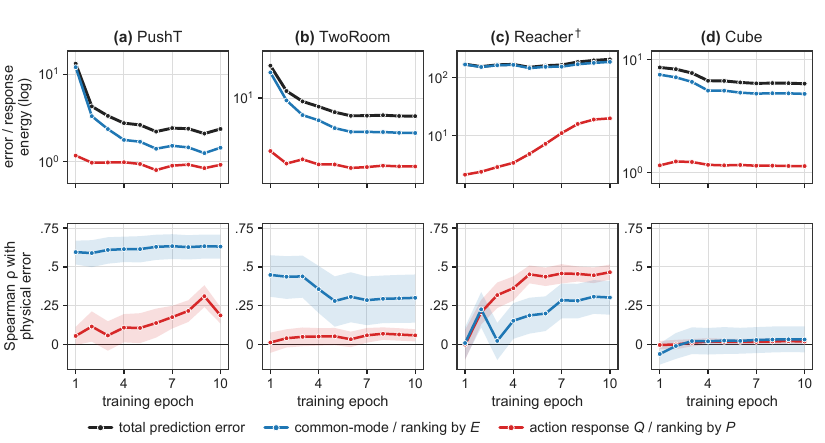}
    \caption{\textbf{Along training, error falls mainly through the center (common-mode) term while predictor ranking lags.} Top: total prediction error and its two exact components, relative to true response energy (log scale). Bottom: rank correlation with physical error for the predicted-outcome cost $P$ and the encoded-outcome cost $E$; bands are descriptive 95\% start-bootstrap intervals. The ten checkpoints of a run are nested, not independent runs. $\dagger$: Reacher's encoded quantities use natively rendered endpoints and are not comparable with the LeWM Reacher row.}
    \label{fig:trajectory}
\end{figure}

Although total error and $Q$ fall on PushT and TwoRoom, $\rho_P$ improves resolvably only on PushT, from .056 to .188 (its interval, [.0003, .262], barely excludes zero), against $\rho_E$ of .595 to .631. On Cube there is no resolved change in $Q$ or $\rho_P$; $\rho_E$ moves from $-.060$ to $.033$ and stays uninformative. The change in the $E-P$ gap is unresolved on every task. Epoch-to-epoch variation of $\rho_P$ is large (PushT .312 at epoch 9 and .188 at epoch 10), candidates were generated by the final-epoch planner and are off-policy for earlier epochs, and one training run per task is available, so the curves describe these runs rather than training in general.

%% file: extended_tables_appendix.tex
\section{Extended result tables}
\label{app:tables}

These tables report the complete model--task results underlying the main-text comparisons, followed by the action-block coverage control.

\begin{table}[!htbp]
\caption{Same-start response geometry at H5 for four further model families (64 starts, $K=4$). Response share is the action-relative fraction of total latent MSE; $Q$ is the ratio-of-sums normalized response error with a descriptive 95\% start-bootstrap interval; visibility is the within-start share of latent variance. PreJEPA and DINO-WM rows are conditional on zero placeholders for extra context that the branch banks did not record. Values are within-model and do not rank architectures.}
\label{tab:geometry}
\centering
\small
\setlength{\tabcolsep}{5pt}
\begin{tabular}{@{}llccccccc@{}}
\toprule
 & & \textbf{Response} & \multicolumn{2}{c}{\textbf{Response error $Q$}} & \textbf{Gain} & \textbf{Cosine} & \multicolumn{2}{c}{\textbf{Branch visibility}} \\
\cmidrule(lr){4-5}\cmidrule(l){8-9}
Model & Task & share & est. & 95\% CI & $g$ & $c$ & true & predicted \\
\midrule
PLDM & PushT & .690 & .665 & \ci{.569}{.769} & .894 & .634 & .612 & .588 \\
 & TwoRoom & .691 & .067 & \ci{.041}{.099} & 1.005 & .967 & .673 & .679 \\
 & Reacher & .755 & .342 & \ci{.288}{.401} & 1.004 & .830 & .744 & .745 \\
 & Cube & .741 & .770 & \ci{.686}{.854} & .832 & .554 & .740 & .743 \\
\addlinespace
PreJEPA & PushT & .489 & 1.348 & \ci{1.292}{1.411} & .707 & .107 & .354 & .223 \\
 & TwoRoom & .634 & 1.145 & \ci{1.021}{1.278} & 1.007 & .432 & .503 & .621 \\
 & Reacher & .717 & .903 & \ci{.860}{.951} & .780 & .452 & .738 & .712 \\
 & Cube & .681 & .924 & \ci{.864}{.988} & .844 & .467 & .669 & .723 \\
\addlinespace
DreamerV3 & PushT & .573 & 1.339 & \ci{1.263}{1.415} & 1.003 & .332 & .466 & .429 \\
 & TwoRoom & .714 & .252 & \ci{.220}{.288} & .987 & .873 & .597 & .583 \\
 & Reacher & .758 & .518 & \ci{.462}{.578} & .997 & .741 & .737 & .732 \\
 & Cube & .728 & .576 & \ci{.539}{.614} & 1.000 & .712 & .708 & .697 \\
\addlinespace
DINO-WM & PushT & .652 & 1.554 & \ci{1.463}{1.658} & .821 & .073 & .356 & .257 \\
 & TwoRoom & .667 & 1.495 & \ci{1.348}{1.648} & 1.048 & .288 & .568 & .680 \\
 & Reacher & .745 & .978 & \ci{.902}{1.054} & .977 & .500 & .742 & .741 \\
 & Cube & .686 & 1.066 & \ci{.977}{1.160} & .961 & .446 & .667 & .744 \\
\bottomrule
\end{tabular}
\end{table}

\begin{table}[!htbp]
\caption{Native CEM success and the transition-minus-base contrast ($\Delta$). Base, transition, and branch average three trained seeds; initial is one checkpoint. Printed intervals are 12-comparison family-wise and all include zero. The distinct PushT base-minus-initial contrast is $-6.641$ percentage points [${-11.979},{-1.432}$] under the same family; it measures shared fine-tuning rather than an added response loss.}
\label{tab:cem}
\centering
\small
\setlength{\tabcolsep}{5pt}
\begin{tabular}{@{}lcccccc@{}}
\toprule
\textbf{Task} & \textbf{Initial} & \textbf{Base} & \textbf{Transition} & \textbf{Branch} & \textbf{$\Delta$} & \textbf{Family-wise CI} \\
\midrule
PushT & 89.06 & 82.42 & 83.07 & 82.42 & 0.651 & \ci{$-$2.344}{3.646} \\
TwoRoom & 85.94 & 86.20 & 85.55 & 86.33 & $-$0.651 & \ci{$-$1.953}{0.521} \\
Reacher & 58.98 & 58.46 & 56.12 & 59.51 & $-$2.344 & \ci{$-$8.073}{3.516} \\
Cube & 63.67 & 64.58 & 65.76 & 64.58 & 1.172 & \ci{$-$1.823}{4.297} \\
\bottomrule
\end{tabular}
\end{table}

\begin{table}[!htbp]
\caption{Matched-candidate contrast: predicted-outcome cost $P$ versus encoded-realized-outcome cost $E$ on identical candidates (64 starts, 300 candidates, top-30 elites). Gains are oriented so that positive favors $E$; elite error is in task-specific units. Bold marks a family-wise interval that excludes zero: eight comparisons for LeWM and separate prespecified 32-comparison families for feedforward and recurrent cross-model protocols.}
\label{tab:matched}
\centering
\small
\setlength{\tabcolsep}{5pt}
\begin{tabular}{@{}llcccccc@{}}
\toprule
 & & \multicolumn{2}{c}{\textbf{Rank corr.\ $\rho$}} & \multicolumn{2}{c}{\textbf{Rank gain $E-P$}} & \multicolumn{2}{c}{\textbf{Elite-error reduction $P-E$}} \\
\cmidrule(lr){3-4}\cmidrule(lr){5-6}\cmidrule(l){7-8}
Model & Task & $P$ & $E$ & est. & family-wise CI & est. & family-wise CI \\
\midrule
LeWM & PushT & .193 & .631 & \sig{.438} & \ci{.319}{.556} & \sig{.175} & \ci{.110}{.248} \\
 & TwoRoom & .060 & .300 & \sig{.241} & \ci{.018}{.448} & .028 & \ci{$-$.037}{.086} \\
 & Reacher & .464 & .975 & \sig{.511} & \ci{.449}{.574} & \sig{1.337} & \ci{1.025}{1.681} \\
 & Cube & .020 & .033 & .013 & \ci{$-$.095}{.122} & .030 & \ci{$-$.103}{.202} \\
\midrule
PLDM & PushT & .267 & .549 & \sig{.282} & \ci{.165}{.409} & \sig{.108} & \ci{.056}{.180} \\
 & TwoRoom & .093 & .322 & .229 & \ci{$-$.050}{.486} & .049 & \ci{$-$.022}{.110} \\
 & Reacher & .444 & .737 & \sig{.293} & \ci{.123}{.449} & .474 & \ci{$-$.179}{1.117} \\
 & Cube & .004 & $-$.052 & $-$.056 & \ci{$-$.165}{.047} & $-$.082 & \ci{$-$.186}{.036} \\
\addlinespace
PreJEPA & PushT & .097 & .516 & \sig{.419} & \ci{.240}{.613} & \sig{.195} & \ci{.098}{.343} \\
 & TwoRoom & .080 & .642 & \sig{.562} & \ci{.401}{.692} & \sig{.108} & \ci{.069}{.143} \\
 & Reacher & .518 & .793 & \sig{.275} & \ci{.167}{.409} & \sig{.741} & \ci{.394}{1.189} \\
 & Cube & .035 & $-$.001 & $-$.036 & \ci{$-$.164}{.090} & .008 & \ci{$-$.141}{.192} \\
\addlinespace
DreamerV3 & PushT & .017 & .293 & \sig{.276} & \ci{.162}{.396} & \sig{.149} & \ci{.074}{.238} \\
 & TwoRoom & .137 & .700 & \sig{.563} & \ci{.411}{.698} & \sig{.123} & \ci{.085}{.167} \\
 & Reacher & .270 & .740 & \sig{.470} & \ci{.324}{.625} & \sig{1.355} & \ci{.907}{1.850} \\
 & Cube & .013 & $-$.042 & $-$.055 & \ci{$-$.182}{.067} & $-$.037 & \ci{$-$.179}{.126} \\
\addlinespace
DINO-WM & PushT & .278 & .564 & \sig{.286} & \ci{.163}{.421} & \sig{.119} & \ci{.049}{.199} \\
 & TwoRoom & .117 & .707 & \sig{.590} & \ci{.428}{.732} & \sig{.125} & \ci{.076}{.174} \\
 & Reacher & .633 & .700 & .068 & \ci{$-$.032}{.164} & $-$.145 & \ci{$-$.398}{.089} \\
 & Cube & .033 & $-$.004 & $-$.036 & \ci{$-$.147}{.071} & .006 & \ci{$-$.142}{.175} \\
\bottomrule
\end{tabular}
\end{table}

\begin{table}[!htbp]
\caption{Action-use control at the final checkpoint: relative increase of one-step latent prediction loss (loss increase divided by true-action loss) when the action is masked or shuffled across 512 fixed probe windows. Every entry is positive; magnitudes are within-model only.}
\label{tab:actionuse}
\centering
\small
\setlength{\tabcolsep}{5pt}
\begin{tabular}{@{}lcccccccc@{}}
\toprule
 & \multicolumn{2}{c}{\textbf{PushT}} & \multicolumn{2}{c}{\textbf{TwoRoom}} & \multicolumn{2}{c}{\textbf{Reacher}} & \multicolumn{2}{c}{\textbf{Cube}} \\
\cmidrule(lr){2-3}\cmidrule(lr){4-5}\cmidrule(lr){6-7}\cmidrule(l){8-9}
Model & mask & shuffle & mask & shuffle & mask & shuffle & mask & shuffle \\
\midrule
LeWM & 30.1 & 51.7 & 195 & 229 & 17.2 & 28.9 & 125 & 193 \\
PLDM & 47.9 & 74.4 & 115 & 137 & 12.2 & 19.4 & 106 & 140 \\
PreJEPA & 0.79 & 1.07 & 3.18 & 4.64 & 1.05 & 1.61 & 1.60 & 2.01 \\
DreamerV3 & 0.20 & 0.34 & 4.60 & 6.53 & 1.61 & 2.38 & 4.34 & 4.96 \\
DINO-WM & 4.17 & 5.47 & 16.4 & 23.6 & 10.6 & 14.3 & 5.78 & 7.23 \\
\bottomrule
\end{tabular}
\end{table}

\begin{table}[!htbp]
\caption{Action-block coverage control. Top: varying-action base minus repeated-action base at H5; $Q$ intervals use the 80-comparison family, whereas cosine changes are descriptive three-seed means. Bottom: varying-minus-repeated native-CEM success in percentage points for each objective, with 24-comparison family-wise intervals over 256 paired source episodes per task.}
\label{tab:coverage-control}
\centering\small
\setlength{\tabcolsep}{5pt}
\begin{tabular}{@{}lrrr@{}}
\toprule
Task & $\Delta Q$ & Family-wise interval & $\Delta$ cosine \\
\midrule
PushT & $-.0263$ & \ci{$-.0485$}{$-.0055$} & $+.0323$ \\
TwoRoom & $-.0070$ & \ci{$-.0541$}{.0314} & $+.0035$ \\
Reacher & $-.0645$ & \ci{$-.0748$}{$-.0545$} & $+.0522$ \\
Cube & $-.0402$ & \ci{$-.0531$}{$-.0279$} & $+.0217$ \\
\bottomrule
\end{tabular}

\medskip
\begin{tabular}{@{}lrrrr@{}}
\toprule
 & \multicolumn{2}{c}{Base objective} & \multicolumn{2}{c}{Transition objective} \\
\cmidrule(lr){2-3}\cmidrule(l){4-5}
Task & Change & Family-wise interval & Change & Family-wise interval \\
\midrule
PushT & $.39$ & \ci{$-3.52$}{4.56} & $-.65$ & \ci{$-4.69$}{3.65} \\
TwoRoom & $1.04$ & \ci{$-.78$}{3.26} & $1.56$ & \ci{$-.52$}{4.17} \\
Reacher & $-4.56$ & \ci{$-11.72$}{2.60} & $-3.78$ & \ci{$-11.07$}{3.26} \\
Cube & $.52$ & \ci{$-3.52$}{4.81} & $2.21$ & \ci{$-1.95$}{6.51} \\
\bottomrule
\end{tabular}
\end{table}

Varying the action blocks improves offline $Q$ on PushT, Reacher and Cube.
All eight native-CEM intervals include zero, leaving the corresponding incremental closed-loop benefits unresolved.
The cosine changes describe the same data intervention; they do not isolate a causal effect of action coverage on direction.

%% file: response_priority_appendix.tex
\section{Matched-candidate repair studies}
\label{app:repair-studies}

These studies score fixed candidate pools with the $P/C/S/E$ endpoint sets of
Section~\ref{sec:matched}. The exploratory pools
(Appendix~\ref{app:repair-priority}) motivated the question; the new-start
confirmation (Appendix~\ref{app:repair-priority-confirm}) supplies the
reported intervals; and a LeWM first-replan study
(Appendix~\ref{app:native-operator}) executes the mean action of oracle-selected
elites instead of scoring individual candidates.

\subsection{Exploratory pools (64 starts per task)}
\label{app:repair-priority}

\paragraph{Scope and estimator.}
This extends the completed P/E comparison to PLDM, PreJEPA, DINO-WM and
DreamerV3 on four tasks. Each task has 64 start--goal pairs from 64 distinct
evaluation episodes, reused across architectures: 256 pairs, not 1,024
independent physical starts. They are distinct evaluation episodes, which does
not rule out overlap with pretraining data, and they form an existing
exploratory sample rather than a new blind one.

All models score the same 300 candidates from the final iteration of the
LeWM CEM run and the same recorded physical outcomes, rather than each
architecture's own optimized proposal distribution or a new CEM execution.
Feedforward predictions use the maximum available real causal history; DINO
patch coordinates are not pooled. Dreamer $E$ is its recurrent posterior from
recorded observations and actually executed actions, with the expert-path
posterior goal, rather than a stateless endpoint image embedding. Candidates that terminate
early are handled as in the LeWM P/E protocol (Appendix~\ref{app:protocol}).

For each start, with means $\bar p,\bar z$, modes are
\begin{equation}
P_j=p_j,\quad C_j=p_j-\bar p+\bar z,\quad
S_j=z_j-\bar z+\bar p,\quad E_j=z_j.
\end{equation}
$C$ preserves every pairwise predicted branch distance while correcting the
center. $S$ retains the predicted center while correcting all centered
responses. Their oracle information budgets differ: $C$ imports one true
pool-mean vector, whereas $S$ imports the candidate-wise true centered
responses. The $S$--$C$ contrast therefore tests decision sensitivity to
ideal component replacement, not which component is cheaper to learn or
repair at equal information cost. Squared Euclidean cost uses every coordinate
the model scores natively, computed in double precision. Spearman uses midranks,
with constant ranks assigned zero; elites are the top 30 by stable
candidate-index tie breaking. Elite physical error averages the selected
candidates, not their action mean.

The six paired contrasts per configuration are $C-P$, $S-P$ and $C-S$, each
for correlation gain and elite-error reduction; positive favors the first
mode. Intervals are adjusted over all 96 comparisons
(Table~\ref{tab:resampling-settings}). All six contrasts and all sixteen
configurations were specified before the C/S computation and are not combined
retrospectively with earlier experiments. Joint support below is the
conjunction of two prespecified comparisons, not a new test.

\input{artifacts/priority_center_means_table_v2.tex}

\paragraph{Repair priority differs from the error budget.}
Center error exceeds half the aggregate squared error in 14/16 settings.
Nevertheless, $S$ has better mean correlation and elite quality than $C$ in
all sixteen. Direct $S>C$ is supported on both metrics in seven settings across
PreJEPA, DINO-WM and DreamerV3; PLDM has no jointly resolved setting. Six of
those seven have center-majority error, so $C$ leaves less total squared error
while $S$ makes better choices. The MSE ordering follows the exact
decomposition; it is descriptive, not a separately tested MSE contrast.
The sixteen directional point estimates are not sixteen independent model
replications or a newly selected cross-case sign test.

Relative to $P$, $S$ improves both metrics in eight settings and $C$ in one;
these counts do not replace the direct comparison. On DINO-WM/Reacher, $C$
removes 92.44\% of latent squared error and $S$ removes 7.56\%, yet mean
correlation is .635 for $C$ versus .836 for $S$, and elite error 1.406 versus
.906. Neither C/S gains nor MSE fractions are additive causal responsibilities,
and $E$ is not a physical-performance upper bound. Response replacement changes
amplitude and normalized structure simultaneously, so this experiment alone
does not isolate amplitude as the cause.

\input{artifacts/priority_center_rho_table_v2.tex}
\input{artifacts/priority_center_elite_table_v2.tex}

\paragraph{Why a large common error need not dominate selection.}
Writing $p_j=\bar p+\hat r_j$ and goal $z_g$ yields the elementary identity
\begin{equation}
\|p_i-z_g\|^2-\|p_j-z_g\|^2
=\|\hat r_i\|^2-\|\hat r_j\|^2
 +2\langle\bar p-z_g,\hat r_i-\hat r_j\rangle.
\end{equation}
The shared scalar $\|\bar p-z_g\|^2$ cancels, but the center still affects
ranking through its interaction with action differences. Center error is
therefore not automatically harmless, and its squared magnitude alone does not
measure its decision impact. The identity constrains how center error can act
on selection; it does not describe how training produces either component.

An exploratory, post-hoc recomputation on all 16 new-start confirmation
cells measured this center projection on their saved 300-candidate pools.
The six jointly supported inversion cells have normalized projection
medians between $.061$ and $.414$, overlapping the ten other cells. Thus
near-orthogonality of center error to predicted candidate differences does
not identify which cells favor $S$; the identity gives a geometric
possibility rather than a cell-level mechanism. No inferential claim is
attached to this diagnostic.

\paragraph{Numerical checks.}
Independent CPU audits recompute all costs in double precision and every
bootstrap sample; using the original single-precision costs changes no
status and no elite membership.

A few starts have constant physical costs (two in PushT, one in Cube) and are
retained. On average, 90.6\% of Cube candidate pairs have distinct physical
costs, and selecting the physically best 30 improves on random selection by
.370, so these pools are not degenerate. Appendix~\ref{app:repair-priority-confirm}
tests the exploratory pattern on new starts.

\subsection{New-start confirmation}
\label{app:repair-priority-confirm}

\paragraph{Pre-specified scope.}
The exploratory result above selected the scientific question; it did not
provide the following intervals. Before new forward evaluation, we froze
256 new physical starts per task, excluding the earlier exploratory starts,
four fixed checkpoints (PLDM, PreJEPA, DINO-WM, DreamerV3), and the same
300 LeWM-proposed candidate sequences and physical outcomes at each start.
The models have architecture-specific latent spaces and goals, so no latent
MSE is compared across models. The physical starts are shared across models
within a task, not 1,024 independent physical starts; pretraining-episode
exclusion is not established. The same real-history and recurrent-posterior
conventions as above are retained. This is a fixed-pool scorer comparison,
not a new CEM search or execution of elite-mean actions.

Besides $P,C,S,E$ defined above, write centered predicted and encoded
candidate responses as $R_p=(p_j-\bar p)_j$ and $R_e=(z_j-\bar z)_j$,
with Frobenius norms $A_p,A_e$. The amplitude-only oracle (AMP) replaces
$R_p$ by $(A_e/A_p)R_p$, keeping its normalized configuration.
The structural oracle (SHAPE) replaces it by $(A_p/A_e)R_e$, keeping predicted
total response energy. Both keep the predicted center; if the response being
rescaled has zero energy, the replacement is undefined and the prediction is
left unchanged. SHAPE changes orientation, relative candidate
lengths and branch-label configuration jointly. It has many more oracle
degrees of freedom than AMP, so their comparison diagnoses which information
matters rather than comparing equal-capacity trainable methods.

\begin{figure}[!htbp]
    \centering
    \includegraphics[width=\linewidth]{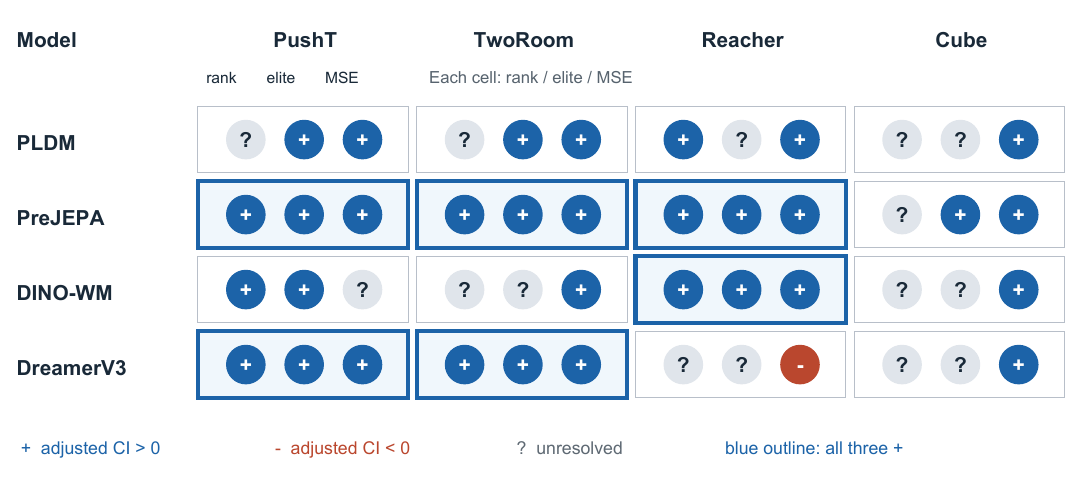}
    \caption{\textbf{The 16-cell new-start confirmation.} Each
    model--task cell shows signs of the 80-comparison family-wise intervals for
    $\rho_S-\rho_C$ (rank), $q_C-q_S$ (top-30 elite improvement), and
    $\mathrm{MSE}_S-\mathrm{MSE}_C$ (more residual MSE under $S$), in that order.
    $+$ means an interval wholly above zero, $-$ wholly below, and $?$ unresolved;
    an outline marks all three positive (six cells).
    Models share 256 physical starts within each task and the same 300
    LeWM-proposed candidates per start. Physical elite-error units are
    task-specific; exact effects and intervals for all cells are in
    Tables~\ref{tab:repair-priority-confirm-status}--\ref{tab:repair-priority-confirm-shape}.}
    \label{fig:repair-map}
\end{figure}

\paragraph{How to read the sign map.}
The five frozen per-cell contrasts are
$\rho_S-\rho_C$, elite-error reduction $q_C-q_S$
(where $q$ is fixed-pool top-30 physical error),
per-coordinate residual $\mathrm{MSE}_S-\mathrm{MSE}_C$,
$\rho_{\rm SHAPE}-\rho_{\rm AMP}$ and
$q_{\rm AMP}-q_{\rm SHAPE}$. All 80 contrasts share one prespecified
family-wise two-sided paired-start bootstrap
(Table~\ref{tab:resampling-settings}).
Figure~\ref{fig:repair-map} and Table~\ref{tab:repair-priority-confirm-status}
are sign maps, \emph{not} five performance gains: $+$ favors $S$ in the first
two decision columns, but means \emph{higher residual latent MSE for $S$} in
the third. Thus $+++$ across the first three columns marks the MSE--decision
inversion, not three improvements. $++$ in the last two columns favors SHAPE
over AMP. We retain every cell without pooling and do not treat unresolved
intervals as equivalence.

\input{artifacts/repair_priority_confirmation_status_v2.tex}

\paragraph{Results across the 16 cells.}
The joint repair-priority reversal holds in six of sixteen cells:
PreJEPA on PushT, TwoRoom and Reacher; DINO-WM on Reacher; DreamerV3
on PushT and TwoRoom. No PLDM cell qualifies. DreamerV3/Reacher is
an explicit counterexample to universal $S$-better-with-worse-MSE:
its residual-MSE contrast is $-.00858$
\ci{$-.01014$}{$-.00691$}, while its direct decision contrasts
are unresolved. The SHAPE-over-AMP joint decision advantage holds
in all twelve non-Cube cells across all four architectures and three
tasks, and in none of the four Cube cells; Cube is unresolved on the
joint claim rather than favoring AMP. For PreJEPA/Reacher, SHAPE's rank
gain is .3671 \ci{.3158}{.4205} and its elite-error reduction is 1.0146
\ci{.8491}{1.1970}.

\paragraph{Representation/goal-metric boundary and sensitivity.}
Even the encoded-realized $E$ cost has physical Spearman means from
$-.044$ to $.040$ across the four Cube checkpoints, versus $.214$
to $.758$ in the twelve non-Cube cells. A post-hoc exact-tie
control on the same new bank finds 155 Cube starts with all 300
physical candidate scores distinct; their four $E$ means remain
$-.062$ to $.064$. PushT has 193 all-distinct starts and corresponding
$E$ means $.248$ to $.563$. Exact physical ties therefore do not explain
the Cube result; near-ties, goal-observation mismatch and other metric
failures remain possible. This control is descriptive.

Independent CPU audits of every forward pass reconstruct all 80
contrasts. One Cube candidate endpoint rendered reproducibly differently
despite identical physical replay; re-scoring all four models with the
alternate image changed no primary status, rank permutation or elite set.

\input{artifacts/repair_priority_confirmation_center_v2.tex}
\input{artifacts/repair_priority_confirmation_shape_v2.tex}

%% file: artifacts/priority_center_means_table_v2.tex
\begin{table}[!htbp]
\caption{All-mode means and aggregate center-error fraction $f_c$. Order is $P,C,S,E$ in both metric blocks. These fractions are descriptive, not causal responsibility. Elite means use task-specific physical units.}
\label{tab:priority-center_means}
\centering\scriptsize
\setlength{\tabcolsep}{3pt}
\begin{tabular}{@{}lccccccccc@{}}
\toprule
Model / task & $f_c$ & $\rho_P$ & $\rho_C$ & $\rho_S$ & $\rho_E$ & $q_P$ & $q_C$ & $q_S$ & $q_E$ \\
\midrule
PLDM / PushT & 0.580 & 0.277 & 0.350 & 0.354 & 0.549 & 0.408 & 0.377 & 0.368 & 0.297 \\
PLDM / TwoRoom & 0.668 & 0.094 & 0.094 & 0.234 & 0.322 & 1.689 & 1.702 & 1.652 & 1.640 \\
PLDM / Reacher & 0.654 & 0.718 & 0.738 & 0.842 & 0.737 & 1.142 & 1.175 & 1.093 & 1.592 \\
PLDM / Cube & 0.754 & 0.007 & 0.038 & 0.116 & -0.052 & 1.290 & 1.298 & 1.234 & 1.373 \\
PreJEPA / PushT & 0.829 & 0.116 & 0.182 & 0.503 & 0.516 & 0.502 & 0.427 & 0.321 & 0.308 \\
PreJEPA / TwoRoom & 0.761 & 0.093 & 0.114 & 0.473 & 0.642 & 1.685 & 1.687 & 1.612 & 1.581 \\
PreJEPA / Reacher & 0.897 & 0.513 & 0.616 & 0.859 & 0.793 & 1.833 & 1.555 & 0.886 & 1.112 \\
PreJEPA / Cube & 0.674 & 0.030 & 0.053 & 0.117 & -0.001 & 1.291 & 1.280 & 1.222 & 1.283 \\
DINO-WM / PushT & 0.443 & 0.240 & 0.332 & 0.478 & 0.564 & 0.422 & 0.392 & 0.333 & 0.297 \\
DINO-WM / TwoRoom & 0.703 & 0.117 & 0.150 & 0.262 & 0.707 & 1.687 & 1.703 & 1.664 & 1.562 \\
DINO-WM / Reacher & 0.924 & 0.633 & 0.635 & 0.836 & 0.700 & 1.265 & 1.406 & 0.906 & 1.406 \\
DINO-WM / Cube & 0.612 & 0.028 & 0.052 & 0.152 & -0.004 & 1.292 & 1.272 & 1.207 & 1.289 \\
DreamerV3 / PushT & 0.694 & 0.017 & 0.019 & 0.254 & 0.293 & 0.542 & 0.559 & 0.406 & 0.393 \\
DreamerV3 / TwoRoom & 0.627 & 0.137 & 0.150 & 0.321 & 0.700 & 1.692 & 1.703 & 1.635 & 1.570 \\
DreamerV3 / Reacher & 0.366 & 0.270 & 0.543 & 0.585 & 0.740 & 2.595 & 1.904 & 1.746 & 1.241 \\
DreamerV3 / Cube & 0.656 & 0.013 & 0.030 & 0.106 & -0.042 & 1.303 & 1.297 & 1.236 & 1.340 \\
\bottomrule
\end{tabular}
\end{table}

%% file: artifacts/priority_center_rho_table_v2.tex
\begin{table}[!htbp]
\caption{Center/response localization: physical Spearman, 48 of the 96 contrasts in this comparison family. Positive favors the first mode in each column; negative $C-S$ favors response repair. All 16 configurations are retained.}
\label{tab:priority-center_rho}
\centering\footnotesize
\setlength{\tabcolsep}{3pt}
\begin{tabular}{@{}lccc@{}}
\toprule
Model / task & $C-P$ & $S-P$ & $C-S$ \\
\midrule
PLDM / PushT & 0.0738 \ci{-0.0147}{0.1656} & 0.0769 \ci{-0.0278}{0.1676} & -0.0032 \ci{-0.1100}{0.1069} \\
PLDM / TwoRoom & 0.0000 \ci{-0.0997}{0.0656} & 0.1407 \ci{-0.0782}{0.3582} & -0.1407 \ci{-0.3637}{0.0724} \\
PLDM / Reacher & 0.0207 \ci{-0.0169}{0.0615} & \textbf{0.1248} \ci{0.0380}{0.2060} & \textbf{-0.1041} \ci{-0.1864}{-0.0046} \\
PLDM / Cube & 0.0310 \ci{-0.0000}{0.0641} & \textbf{0.1086} \ci{0.0130}{0.2013} & -0.0776 \ci{-0.1725}{0.0211} \\
PreJEPA / PushT & 0.0664 \ci{-0.0651}{0.1973} & \textbf{0.3870} \ci{0.2087}{0.5717} & \textbf{-0.3206} \ci{-0.4786}{-0.1586} \\
PreJEPA / TwoRoom & 0.0218 \ci{-0.0712}{0.1008} & \textbf{0.3806} \ci{0.1966}{0.5325} & \textbf{-0.3588} \ci{-0.5301}{-0.1774} \\
PreJEPA / Reacher & \textbf{0.1030} \ci{0.0293}{0.1955} & \textbf{0.3466} \ci{0.2357}{0.4785} & \textbf{-0.2436} \ci{-0.3335}{-0.1640} \\
PreJEPA / Cube & 0.0234 \ci{-0.0089}{0.0722} & 0.0872 \ci{-0.0279}{0.1882} & -0.0638 \ci{-0.1558}{0.0486} \\
DINO-WM / PushT & 0.0923 \ci{-0.0028}{0.1922} & \textbf{0.2379} \ci{0.1130}{0.3633} & \textbf{-0.1457} \ci{-0.2607}{-0.0341} \\
DINO-WM / TwoRoom & 0.0333 \ci{-0.0803}{0.1195} & 0.1449 \ci{-0.0166}{0.3030} & -0.1116 \ci{-0.2944}{0.0797} \\
DINO-WM / Reacher & 0.0015 \ci{-0.0458}{0.0441} & \textbf{0.2026} \ci{0.1202}{0.2905} & \textbf{-0.2011} \ci{-0.2872}{-0.1199} \\
DINO-WM / Cube & 0.0237 \ci{-0.0006}{0.0520} & \textbf{0.1242} \ci{0.0264}{0.2171} & \textbf{-0.1004} \ci{-0.1920}{-0.0065} \\
DreamerV3 / PushT & 0.0024 \ci{-0.0653}{0.0684} & \textbf{0.2373} \ci{0.1181}{0.3586} & \textbf{-0.2349} \ci{-0.3544}{-0.1235} \\
DreamerV3 / TwoRoom & 0.0132 \ci{-0.0333}{0.0628} & \textbf{0.1844} \ci{0.0592}{0.3329} & \textbf{-0.1712} \ci{-0.3349}{-0.0344} \\
DreamerV3 / Reacher & \textbf{0.2732} \ci{0.1689}{0.3863} & \textbf{0.3149} \ci{0.2097}{0.4242} & -0.0417 \ci{-0.1486}{0.0776} \\
DreamerV3 / Cube & 0.0170 \ci{-0.0085}{0.0471} & 0.0930 \ci{-0.0319}{0.2092} & -0.0759 \ci{-0.1977}{0.0492} \\
\bottomrule
\end{tabular}
\end{table}

%% file: artifacts/priority_center_elite_table_v2.tex
\begin{table}[!htbp]
\caption{Center/response localization: elite-error reduction, 48 of the 96 contrasts in this comparison family. Positive favors the first mode in each column; negative $C-S$ favors response repair. All 16 configurations are retained.}
\label{tab:priority-center_elite}
\centering\footnotesize
\setlength{\tabcolsep}{3pt}
\begin{tabular}{@{}lccc@{}}
\toprule
Model / task & $C-P$ & $S-P$ & $C-S$ \\
\midrule
PLDM / PushT & \textbf{0.0315} \ci{0.0009}{0.0700} & \textbf{0.0397} \ci{0.0009}{0.0941} & -0.0082 \ci{-0.0554}{0.0337} \\
PLDM / TwoRoom & -0.0121 \ci{-0.0462}{0.0076} & 0.0371 \ci{-0.0229}{0.0929} & -0.0492 \ci{-0.1192}{0.0114} \\
PLDM / Reacher & -0.0334 \ci{-0.1552}{0.0782} & 0.0484 \ci{-0.2161}{0.2478} & -0.0818 \ci{-0.2909}{0.1828} \\
PLDM / Cube & -0.0081 \ci{-0.0995}{0.0595} & 0.0560 \ci{-0.0152}{0.1778} & -0.0641 \ci{-0.1750}{0.0221} \\
PreJEPA / PushT & 0.0755 \ci{-0.0125}{0.2086} & \textbf{0.1813} \ci{0.0845}{0.3330} & \textbf{-0.1058} \ci{-0.1818}{-0.0521} \\
PreJEPA / TwoRoom & -0.0013 \ci{-0.0357}{0.0215} & \textbf{0.0732} \ci{0.0270}{0.1101} & \textbf{-0.0745} \ci{-0.1135}{-0.0354} \\
PreJEPA / Reacher & 0.2780 \ci{-0.0758}{0.6443} & \textbf{0.9474} \ci{0.6157}{1.3573} & \textbf{-0.6694} \ci{-0.9472}{-0.4435} \\
PreJEPA / Cube & 0.0113 \ci{-0.0571}{0.0895} & 0.0695 \ci{-0.0241}{0.2058} & -0.0582 \ci{-0.1707}{0.0242} \\
DINO-WM / PushT & 0.0302 \ci{-0.0315}{0.0844} & \textbf{0.0890} \ci{0.0375}{0.1512} & \textbf{-0.0588} \ci{-0.1084}{-0.0096} \\
DINO-WM / TwoRoom & -0.0159 \ci{-0.0885}{0.0197} & 0.0225 \ci{-0.0262}{0.0685} & -0.0384 \ci{-0.1013}{0.0146} \\
DINO-WM / Reacher & -0.1419 \ci{-0.3440}{0.0124} & \textbf{0.3586} \ci{0.1904}{0.5220} & \textbf{-0.5005} \ci{-0.7388}{-0.2978} \\
DINO-WM / Cube & 0.0196 \ci{-0.0216}{0.0736} & 0.0851 \ci{-0.0142}{0.2117} & -0.0655 \ci{-0.1533}{0.0194} \\
DreamerV3 / PushT & -0.0168 \ci{-0.0732}{0.0376} & \textbf{0.1356} \ci{0.0597}{0.2284} & \textbf{-0.1524} \ci{-0.2554}{-0.0711} \\
DreamerV3 / TwoRoom & -0.0105 \ci{-0.0330}{0.0068} & \textbf{0.0575} \ci{0.0156}{0.1311} & \textbf{-0.0680} \ci{-0.1539}{-0.0214} \\
DreamerV3 / Reacher & \textbf{0.6906} \ci{0.4329}{0.9769} & \textbf{0.8487} \ci{0.5536}{1.1606} & -0.1580 \ci{-0.4098}{0.0941} \\
DreamerV3 / Cube & 0.0063 \ci{-0.0518}{0.0537} & 0.0673 \ci{-0.0290}{0.2083} & -0.0611 \ci{-0.1942}{0.0214} \\
\bottomrule
\end{tabular}
\end{table}

%% file: artifacts/repair_priority_confirmation_status_v2.tex
\begin{table}[!htbp]
\caption{New-start confirmation: 256 starts per task, disjoint from the exploratory starts, and 300 identical LeWM-proposed candidates per start; overlap with pretraining episodes is not ruled out. Entries are the signs of the five preregistered contrasts, with intervals adjusted over all 80 comparisons, in order: $\rho_S-\rho_C$, $q_C-q_S$, $\mathrm{MSE}_S-\mathrm{MSE}_C$, $\rho_{\rm SHAPE}-\rho_{\rm AMP}$, $q_{\rm AMP}-q_{\rm SHAPE}$. $+$ denotes an adjusted interval wholly above zero, $-$ wholly below, and ? unresolved. Cells are not pooled. All repairs are oracles that use realized outcomes, not deployable methods or closed-loop results.}
\label{tab:repair-priority-confirm-status}
\centering\scriptsize
\setlength{\tabcolsep}{3pt}
\begin{tabular}{@{}lccccc@{}}
\toprule
Model / task & S-C rank & S-C elite & S-C MSE & Shape-Amp rank & Shape-Amp elite \\
\midrule
PLDM / PushT & ? & + & + & + & + \\
PLDM / TwoRoom & ? & + & + & + & + \\
PLDM / Reacher & + & ? & + & + & + \\
PLDM / Cube & ? & ? & + & ? & ? \\
PreJEPA / PushT & + & + & + & + & + \\
PreJEPA / TwoRoom & + & + & + & + & + \\
PreJEPA / Reacher & + & + & + & + & + \\
PreJEPA / Cube & ? & + & + & ? & + \\
DINO-WM / PushT & + & + & ? & + & + \\
DINO-WM / TwoRoom & ? & ? & + & + & + \\
DINO-WM / Reacher & + & + & + & + & + \\
DINO-WM / Cube & ? & ? & + & ? & ? \\
DreamerV3 / PushT & + & + & + & + & + \\
DreamerV3 / TwoRoom & + & + & + & + & + \\
DreamerV3 / Reacher & ? & ? & - & + & + \\
DreamerV3 / Cube & ? & ? & + & ? & ? \\
\bottomrule
\end{tabular}
\end{table}

%% file: artifacts/repair_priority_confirmation_center_v2.tex
\begin{table}[!htbp]
\caption{Center-versus-response repair on the new starts: all 48 contrasts, with intervals adjusted over all 80 comparisons. Positive rank/elite values favor response repair S; positive residual-MSE values mean S leaves more latent error than center repair C. Bold indicates a family-wise interval excluding zero. Elite units are task-specific physical errors.}
\label{tab:repair-priority-confirm-center}
\centering\scriptsize
\setlength{\tabcolsep}{3pt}
\begin{tabular}{@{}lccc@{}}
\toprule
Model / task & $\rho_S-\rho_C$ & $q_C-q_S$ & $\mathrm{MSE}_S-\mathrm{MSE}_C$ \\
\midrule
PLDM / PushT & 0.0567 \ci{-0.0049}{0.1165} & \textbf{0.0421} \ci{0.0044}{0.0914} & \textbf{0.0735} \ci{0.0316}{0.1235} \\
PLDM / TwoRoom & 0.0791 \ci{-0.0140}{0.1714} & \textbf{0.0291} \ci{0.0023}{0.0556} & \textbf{0.3166} \ci{0.1868}{0.4504} \\
PLDM / Reacher & \textbf{0.1336} \ci{0.0968}{0.1709} & 0.0515 \ci{-0.0389}{0.1318} & \textbf{0.1164} \ci{0.0692}{0.1683} \\
PLDM / Cube & 0.0333 \ci{-0.0119}{0.0788} & 0.0294 \ci{-0.0240}{0.0919} & \textbf{0.6783} \ci{0.4983}{0.8621} \\
PreJEPA / PushT & \textbf{0.2436} \ci{0.1668}{0.3186} & \textbf{0.0982} \ci{0.0386}{0.1662} & \textbf{0.3787} \ci{0.3504}{0.4081} \\
PreJEPA / TwoRoom & \textbf{0.2626} \ci{0.1816}{0.3437} & \textbf{0.0647} \ci{0.0355}{0.0989} & \textbf{0.2452} \ci{0.1906}{0.3052} \\
PreJEPA / Reacher & \textbf{0.2900} \ci{0.2442}{0.3379} & \textbf{0.7475} \ci{0.6279}{0.8742} & \textbf{1.0953} \ci{1.0732}{1.1188} \\
PreJEPA / Cube & 0.0303 \ci{-0.0214}{0.0826} & \textbf{0.0692} \ci{0.0005}{0.1680} & \textbf{0.4346} \ci{0.3126}{0.5644} \\
DINO-WM / PushT & \textbf{0.1295} \ci{0.0615}{0.1995} & \textbf{0.0814} \ci{0.0367}{0.1482} & -0.0088 \ci{-0.0287}{0.0110} \\
DINO-WM / TwoRoom & 0.0578 \ci{-0.0375}{0.1559} & 0.0183 \ci{-0.0178}{0.0566} & \textbf{0.1023} \ci{0.0768}{0.1353} \\
DINO-WM / Reacher & \textbf{0.2411} \ci{0.1983}{0.2848} & \textbf{0.4916} \ci{0.3995}{0.5930} & \textbf{1.6051} \ci{1.5822}{1.6288} \\
DINO-WM / Cube & 0.0440 \ci{-0.0068}{0.0950} & 0.0462 \ci{-0.0247}{0.1394} & \textbf{0.3077} \ci{0.1727}{0.4522} \\
DreamerV3 / PushT & \textbf{0.1852} \ci{0.1264}{0.2428} & \textbf{0.1201} \ci{0.0720}{0.1880} & \textbf{0.0283} \ci{0.0240}{0.0329} \\
DreamerV3 / TwoRoom & \textbf{0.1357} \ci{0.0544}{0.2190} & \textbf{0.0410} \ci{0.0061}{0.0825} & \textbf{0.0066} \ci{0.0043}{0.0092} \\
DreamerV3 / Reacher & 0.0486 \ci{-0.0103}{0.1075} & 0.1144 \ci{-0.0196}{0.2439} & \textbf{-0.0086} \ci{-0.0101}{-0.0069} \\
DreamerV3 / Cube & 0.0343 \ci{-0.0231}{0.0931} & 0.0176 \ci{-0.0599}{0.0812} & \textbf{0.0214} \ci{0.0143}{0.0285} \\
\bottomrule
\end{tabular}
\end{table}

%% file: artifacts/repair_priority_confirmation_shape_v2.tex
\begin{table}[!htbp]
\caption{Response-configuration versus amplitude-only oracle repair on the new starts: all 32 contrasts, with intervals adjusted over all 80 comparisons. SHAPE keeps predicted total response energy while substituting the true normalized branch configuration; AMP rescales the predicted response to the true total energy. Positive favors SHAPE. No Cube cell supports both decision contrasts; individual contrasts may resolve.}
\label{tab:repair-priority-confirm-shape}
\centering\scriptsize
\setlength{\tabcolsep}{3pt}
\begin{tabular}{@{}lcc@{}}
\toprule
Model / task & $\rho_{\rm SHAPE}-\rho_{\rm AMP}$ & $q_{\rm AMP}-q_{\rm SHAPE}$ \\
\midrule
PLDM / PushT & \textbf{0.1367} \ci{0.0838}{0.1930} & \textbf{0.0817} \ci{0.0413}{0.1382} \\
PLDM / TwoRoom & \textbf{0.0917} \ci{0.0120}{0.1709} & \textbf{0.0280} \ci{0.0009}{0.0560} \\
PLDM / Reacher & \textbf{0.1890} \ci{0.1466}{0.2329} & \textbf{0.1582} \ci{0.0577}{0.2508} \\
PLDM / Cube & 0.0365 \ci{-0.0088}{0.0819} & 0.0173 \ci{-0.0380}{0.0847} \\
PreJEPA / PushT & \textbf{0.3159} \ci{0.2471}{0.3882} & \textbf{0.1633} \ci{0.1084}{0.2338} \\
PreJEPA / TwoRoom & \textbf{0.2987} \ci{0.2301}{0.3664} & \textbf{0.0745} \ci{0.0504}{0.1010} \\
PreJEPA / Reacher & \textbf{0.3671} \ci{0.3158}{0.4205} & \textbf{1.0146} \ci{0.8491}{1.1970} \\
PreJEPA / Cube & 0.0396 \ci{-0.0094}{0.0875} & \textbf{0.0995} \ci{0.0142}{0.2241} \\
DINO-WM / PushT & \textbf{0.1881} \ci{0.1261}{0.2532} & \textbf{0.1048} \ci{0.0599}{0.1690} \\
DINO-WM / TwoRoom & \textbf{0.1109} \ci{0.0322}{0.1875} & \textbf{0.0388} \ci{0.0075}{0.0732} \\
DINO-WM / Reacher & \textbf{0.2418} \ci{0.1947}{0.2895} & \textbf{0.4507} \ci{0.3692}{0.5566} \\
DINO-WM / Cube & 0.0461 \ci{-0.0023}{0.0936} & 0.0545 \ci{-0.0057}{0.1367} \\
DreamerV3 / PushT & \textbf{0.1781} \ci{0.1180}{0.2381} & \textbf{0.1141} \ci{0.0631}{0.1819} \\
DreamerV3 / TwoRoom & \textbf{0.1178} \ci{0.0469}{0.1859} & \textbf{0.0345} \ci{0.0040}{0.0702} \\
DreamerV3 / Reacher & \textbf{0.3182} \ci{0.2643}{0.3722} & \textbf{0.9000} \ci{0.7336}{1.0778} \\
DreamerV3 / Cube & 0.0270 \ci{-0.0255}{0.0796} & 0.0099 \ci{-0.0573}{0.0782} \\
\bottomrule
\end{tabular}
\end{table}

%% file: native_operator_appendix.tex
\subsection{LeWM first-replan execution}
\label{app:native-operator}

The fixed-candidate confirmation scores individual candidates; it does not
execute the native planner's action. A separate LeWM experiment rescored the
same 300 final-iteration CEM candidates at each of 64 exploratory starts per
task under the $P/C/S/E$ modes, selected the top 30, averaged their
action sequences and physically executed that mean action plan. Its
predeclared endpoint was executed physical error, with three contrasts on
four tasks in a 12-comparison family. PushT supported $C-P$ and $C-S$;
TwoRoom and Cube had no supported contrast. In the original Reacher run the goal
and candidate images were rendered differently, so that cell is not a clean
test; a separate Reacher run with render-matched candidate images and the
dataset goal, analysed in its own four-comparison family, supported $C-P$,
$S-P$, $C-S$ and $E-P$.

To see where the two protocols diverge, we retrospectively reread each
task's audited per-start files without new model, environment or bootstrap
calls. For each mode, the first column below is the mean physical error
of its 30 \emph{individually executed} candidate endpoints. The second is
the physical error of the separately executed \emph{mean action sequence}
of those 30 elites. The same start, candidate pool, goal and proxy are
used within a row. All values are $C-S$ (negative favors center); the
last column is the descriptive executed-minus-fixed change. The Reacher
row comes from the render-matched run.

\begin{table}[h]
\caption{Retrospective first-replan comparison on audited LeWM
oracles ($64$ starts/task, final CEM iteration, 300 candidates, top 30).
Fixed-elite and change columns are descriptive; the executed-mean status
is inherited from the original protocols. Physical-error units differ by task.}
\label{tab:native-operator}
\centering
\small
\setlength{\tabcolsep}{5pt}
\begin{tabular}{@{}lrrrl@{}}
\toprule
Task & Fixed elites & Executed mean & Change & Executed $C-S$ status \\
\midrule
PushT & $-.041220$ & $-.069815$ & $-.028595$ & 12-comparison: supported \\
TwoRoom & $.064196$ & $.036242$ & $-.027954$ & 12-comparison: unresolved \\
Reacher (render-matched) & $-.343188$ & $-.489133$ & $-.145945$ & 4-comparison: supported \\
Cube (full render) & $.014597$ & $-.073012$ & $-.087609$ & 12-comparison: unresolved \\
\bottomrule
\end{tabular}
\end{table}

PushT and render-matched Reacher already favor center in fixed-elite point
estimates, so mean-action execution did not \emph{create} their center
preference. Execution shifts $C-S$ toward center in all four rows, and
Cube changes descriptive sign; with four rows and no frozen interval family,
this pattern does not identify the action-mean operator as a cross-task
mechanism, because elite membership, averaging and nonlinear physical
dynamics change together. The four-family confirmation uses other model
families, 256 new starts per task, shared LeWM-proposed pools and
fixed-candidate metrics, so the two studies are complementary rather than
one combined causal comparison. Both runs passed independent audits.

%% file: objective_appendix.tex
\section{Response shrinkage and objective-function studies}
\label{app:shrinkage}

\subsection{Separating direction from amplitude}

Figure~\ref{fig:shrinkage} uses saved per-start response energies
($\|R\|^2$, $\|\hat R\|^2$, $\langle\hat R,R\rangle$ and $\|\hat R-R\|^2$)
from the six-objective study: 19 checkpoints per task, comprising the initial
model and six objectives at three paired seeds, evaluated on 1,024 held-out
starts, four branches and H1--H20. Gain, cosine, $Q$, direction floor and
amplitude excess are pooled ratios of sums over starts, preserving the
identity $Q=(1-c^2)+(g-c)^2$.

This is a retrospective analysis of the trained models. The four contrasts,
two horizons and 32-comparison correction were fixed before computing their
gain and cosine; PushT $Q$ and center fractions had already been examined.
An independent implementation recomputed the identity and bootstrap intervals.
The initial Reacher checkpoint was trained on dataset frames and evaluated
on simulator-rendered branches, so its gain of 3.8 reflects an observation
shift as well as prediction behavior.

For nonnegative cosine, squared error at fixed direction is minimized by
$g=c$. The teacher-forced initial model maintains amplitude as direction
agreement decreases: on PushT, $g-c$ is .07, .22, .36 and .53 at H1, H5, H10
and H20. After 500 squared-error updates, the H5 gaps are .02, .02, $-$.06
and .06 on PushT, TwoRoom, Reacher and Cube. Table~\ref{tab:shrinkage}
separates the resulting amplitude and direction changes. PushT's long-horizon
gain falls from .71 to .25 with a resolved reduction in amplitude excess
and an unresolved direction-floor change. TwoRoom and Cube instead show
direction improvements. Gain-pinned objectives keep larger responses and
over-expand on Reacher beyond H5 (H20 gain 1.79 and $Q=2.11$).

\begin{table}[!htbp]
\caption{Shrinkage-law contrasts (class means over three seeds; paired start bootstrap, 32-comparison family-wise intervals). Base $-$ initial: change in the direction floor $1-c^2$ and in the amplitude excess $(g-c)^2$ under squared-error fine-tuning. Direction $-$ base: change in gain and in excess when the gain is pinned. Bold marks an interval that excludes zero.}
\label{tab:shrinkage}
\centering
\small
\setlength{\tabcolsep}{3.5pt}
\begin{tabular}{@{}llcccccccc@{}}
\toprule
 & & \multicolumn{4}{c}{\textbf{Base $-$ initial}} & \multicolumn{4}{c}{\textbf{Direction $-$ base}} \\
\cmidrule(lr){3-6}\cmidrule(l){7-10}
 & & \multicolumn{2}{c}{$\Delta$ floor} & \multicolumn{2}{c}{$\Delta$ excess} & \multicolumn{2}{c}{$\Delta$ gain} & \multicolumn{2}{c}{$\Delta$ excess} \\
\cmidrule(lr){3-4}\cmidrule(lr){5-6}\cmidrule(lr){7-8}\cmidrule(l){9-10}
Task & H & est. & CI & est. & CI & est. & CI & est. & CI \\
\midrule
PushT & H5 & $-$.010 & \ci{$-$.033}{.012} & \sig{$-$.050} & \ci{$-$.071}{$-$.032} & \sig{.075} & \ci{.065}{.087} & \sig{.006} & \ci{.001}{.011} \\
PushT & H20 & .007 & \ci{$-$.004}{.019} & \sig{$-$.274} & \ci{$-$.332}{$-$.222} & \sig{.155} & \ci{.137}{.174} & \sig{.040} & \ci{.028}{.054} \\
TwoRoom & H5 & \sig{$-$.020} & \ci{$-$.037}{$-$.007} & \sig{$-$.002} & \ci{$-$.003}{$-$.001} & \sig{.002} & \ci{.001}{.002} & \sig{.000} & \ci{.000}{.000} \\
TwoRoom & H20 & \sig{$-$.246} & \ci{$-$.298}{$-$.196} & \sig{$-$.091} & \ci{$-$.130}{$-$.062} & \sig{.003} & \ci{.001}{.004} & .000 & \ci{$-$.001}{.001} \\
Reacher & H5 & \sig{$-$.375} & \ci{$-$.394}{$-$.357} & \sig{$-$14.026} & \ci{$-$14.891}{$-$13.189} & \sig{.331} & \ci{.323}{.339} & \sig{.058} & \ci{.045}{.070} \\
Reacher & H20 & \sig{$-$.165} & \ci{$-$.180}{$-$.150} & \sig{$-$15.131} & \ci{$-$15.821}{$-$14.461} & \sig{1.382} & \ci{1.349}{1.417} & \sig{1.449} & \ci{1.363}{1.537} \\
Cube & H5 & \sig{$-$.310} & \ci{$-$.341}{$-$.280} & \sig{$-$.057} & \ci{$-$.072}{$-$.045} & \sig{.019} & \ci{.018}{.021} & \sig{.003} & \ci{.002}{.004} \\
Cube & H20 & \sig{$-$.342} & \ci{$-$.371}{$-$.313} & \sig{$-$.375} & \ci{$-$.437}{$-$.316} & \sig{.029} & \ci{.026}{.033} & \sig{.013} & \ci{.010}{.015} \\
\bottomrule
\end{tabular}
\end{table}

\subsection{Response objectives under a matched training budget}

Base minus initial measures the effect of 500 updates on branch data, whereas
direction minus base holds data and update budget fixed and changes the
response objective. The transition objective improves Reacher $Q$ by
8.44\%/6.40\% at H5/H20, with descriptive floor changes of $-$.050/$-$.077
and excess changes of at most .02; the H5 fixed-pool regret improvement is
also resolved. Its PushT and Cube $Q$ gains are .68\% and .35\%, respectively,
and TwoRoom is unresolved. Cosine-only training lowers the floor in every
task while allowing excess amplitude: on Reacher at H20, excess is 4.01
versus .004 for base.

To isolate supervision that separates direction and scale, one additional
arm retains the common-mode squared error and replaces only the centered
response term with
\[
\tfrac12\,\mathrm{mean}_{h=1,\ldots,5}(1-c_h)
+\tfrac12\,\mathrm{mean}_{h=1,\ldots,5}(\log g_h)^2,
\]
where $g_h,c_h$ are pooled over the batch. Initialization, data, optimizer,
batch order and the 500-update budget match base, and the weights are fixed
before training. The primary direction-floor and fixed-pool regret
contrasts at H5/H20 use a 16-comparison correction; gain, excess, $Q$ and MSE
use a separate 40-comparison correction. The prespecified outcome classes
are joint floor/regret improvement, direction-only improvement, harm
(increased regret), and null.

Table~\ref{tab:decomposed} reports every task and horizon. Gain exceeds base
in all eight cells. The floor improves on PushT and Reacher and slightly
on TwoRoom at H5; TwoRoom H20 is unresolved, and Cube's floor increases.
Every regret interval includes zero. On Reacher, gain is .995 at H5 but
1.91 at H20, showing that training-horizon calibration can coexist with
long-horizon over-expansion.

A prespecified PushT replication uses three fresh seeds, 2,000 updates,
and a same-seed, same-budget base. It reproduces floor improvements of
$-$.023 \ci{$-$.039}{$-$.007} at H5 and $-$.014 \ci{$-$.024}{$-$.004}
at H20; regret changes remain unresolved at $-$.029 \ci{$-$.105}{.041}
and $-$.082 \ci{$-$.209}{.045}. All six seed-level floor contrasts lie
between $-$.011 and $-$.025; H5 cosine changes from .458 to .482.
The stopping rule ended this arm before native-CEM evaluation.

\begin{table}[!htbp]
\caption{Direction plus log-amplitude objective minus the squared-error base (class means over three paired seeds; paired start bootstrap). Floor and regret are the primary contrasts (16-comparison family); gain and excess are secondary (40-comparison family). Bold marks an interval that excludes zero.}
\label{tab:decomposed}
\centering
\footnotesize
\setlength{\tabcolsep}{2.2pt}
\begin{tabular}{@{}llccccccccl@{}}
\toprule
 & & \multicolumn{2}{c}{\textbf{$\Delta$ floor}} & \multicolumn{2}{c}{\textbf{$\Delta$ regret}} & \multicolumn{2}{c}{\textbf{$\Delta$ gain}} & \multicolumn{2}{c}{\textbf{$\Delta$ excess}} & \\
\cmidrule(lr){3-4}\cmidrule(lr){5-6}\cmidrule(lr){7-8}\cmidrule(lr){9-10}
Task & H & est. & CI & est. & CI & est. & CI & est. & CI & outcome \\
\midrule
PushT & H5 & \sig{$-$.025} & \ci{$-$.043}{$-$.009} & $-$.075 & \ci{$-$.169}{.017} & \sig{.258} & \ci{.235}{.284} & \sig{.061} & \ci{.041}{.085} & direction only \\
PushT & H20 & \sig{$-$.016} & \ci{$-$.025}{$-$.007} & $-$.137 & \ci{$-$.321}{.039} & \sig{.535} & \ci{.494}{.578} & \sig{.336} & \ci{.280}{.400} & direction only \\
TwoRoom & H5 & \sig{$-$.003} & \ci{$-$.005}{$-$.000} & .000 & \ci{$-$.003}{.005} & \sig{.008} & \ci{.006}{.010} & \sig{.000} & \ci{.000}{.001} & direction only \\
TwoRoom & H20 & $-$.004 & \ci{$-$.018}{.010} & .001 & \ci{$-$.012}{.015} & \sig{.015} & \ci{.010}{.020} & \sig{.002} & \ci{.000}{.005} & null \\
Reacher & H5 & \sig{$-$.110} & \ci{$-$.118}{$-$.102} & $-$.094 & \ci{$-$.217}{.022} & \sig{.437} & \ci{.425}{.448} & \sig{.084} & \ci{.070}{.099} & direction only \\
Reacher & H20 & \sig{$-$.176} & \ci{$-$.193}{$-$.159} & .117 & \ci{$-$.162}{.407} & \sig{1.506} & \ci{1.468}{1.542} & \sig{1.787} & \ci{1.673}{1.911} & direction only \\
Cube & H5 & \sig{.005} & \ci{.002}{.008} & $-$.001 & \ci{$-$.009}{.006} & \sig{.056} & \ci{.053}{.058} & \sig{.011} & \ci{.008}{.014} & null \\
Cube & H20 & \sig{.019} & \ci{.012}{.026} & .013 & \ci{$-$.008}{.036} & \sig{.066} & \ci{.060}{.072} & \sig{.049} & \ci{.040}{.058} & null \\
\bottomrule
\end{tabular}
\end{table}

\subsection{Connection to candidate decisions}

Across the 18 trained checkpoints per task, the corrected analysis does not
show a stronger association of $Q$ than MSE with fixed-pool regret in any
of the eight task--horizon strata. An exploratory comparison of the seven
objective-class means on PushT suggested a more specific reversal:
at H20, the three squared-error objectives had the lowest $Q$ but the
largest regret, while the regret ordering followed cosine more closely.
With seven class means and no interval for that ordering, this observation
served as the hypothesis for the separate preregistered replication
(Appendix~\ref{app:stage-b}).

%% file: stage_b_appendix.tex
\section{Preregistered replication: response and decision ordering}
\label{app:stage-b}

\paragraph{Question and design.}
This replication tests whether the exploratory ordering across trained objectives
reappears on fresh physical starts. Four directional predictions were fixed
before data collection. The study evaluates 19 frozen PushT checkpoints
(initial plus six objectives at three seeds) on 2,048 new starts, four
candidate branches and an independent goal branch, at H5 and H20.

The initial checkpoint is an anchor. The primary contrast is the mean of
nine non-shrinkage checkpoints (NS: direction calibration, the combined
objective and cosine-only supervision) minus nine squared-error checkpoints
(SQ: base, branch geometry and transition-increment supervision).
The prespecified predictions are higher $Q$ but lower fixed-pool regret
for NS at both horizons. The sample size was increased from 1,024 to 2,048
after the exploratory analysis but before collecting the new starts; objective
classes, predicted directions and decision rules stayed fixed. None of the
new starts coincides with a start used in training or earlier evaluation
(checked by complete simulator state).
Paired whole-start bootstrap and its four-comparison correction are specified
in Table~\ref{tab:resampling-settings}.

\begin{table}[!htbp]
\caption{All four primary contrasts of the preregistered replication, NS minus SQ. Intervals are Bonferroni-adjusted over the four prespecified comparisons. Positive $Q$ favors the squared-error class; negative regret favors the non-shrinkage class.}
\label{tab:stage-b}
\centering\small
\setlength{\tabcolsep}{5pt}
\begin{tabular}{@{}llcrrl@{}}
\toprule
Horizon & Quantity & Predicted sign & Estimate & Family-wise interval & Outcome \\
\midrule
H20 & $Q$ & $+$ & $+.0438$ & \ci{.0327}{.0558} & Predicted direction \\
H20 & Regret & $-$ & $-.0635$ & \ci{$-.1273$}{.0007} & Unresolved \\
H5 & $Q$ & $+$ & $-.0219$ & \ci{$-.0285$}{$-.0154$} & Opposite direction \\
H5 & Regret & $-$ & $-.0251$ & \ci{$-.0586$}{.0084} & Unresolved \\
\bottomrule
\end{tabular}
\end{table}

\paragraph{Results and interpretation.}
The H20 response ordering is reproduced, while its regret interval includes
zero. The H5 response contrast has the opposite sign to its prediction.
The preregistered decision rule therefore yields response-only support
(\texttt{RESPONSE\_ONLY}) rather than a confirmed decision reversal.
This result concerns trained objectives on a fixed candidate pool; it
addresses a different intervention from true-future $C/S$ replacement
and a different endpoint from native-CEM success. Checkpoints remain fixed
throughout, and data collection ends at the amended sample size.

%% file: continuation_appendix.tex
\section{Exploratory continuation probes}
\label{app:mechanism}

The main results concern the decomposition and the matched-candidate repairs.
This appendix collects exploratory probes asking whether the direction of a
model's own residual explains how its error propagates through the next
prediction. They use four new physical banks (1,024 starts per task, four
branches) and sixteen fixed LeWM/PLDM checkpoints. Later probes were designed
after earlier results and are reported in that order. The consistent finding
is a condition-associated directional sensitivity that other same-start
directions and real-trajectory controls share, so it is not specific to
predictor error.

\subsection{Directional continuation}
\label{app:continuation}

\paragraph{Setup.}
Let $e$ be the true encoding at H5, $\delta$ an error direction, and $F$ the
model's next prediction step with its history and action held fixed. The
squared directional response is
\begin{equation}
\kappa_\lambda(\delta)=\frac{\|F(e+\lambda\delta)-F(e-\lambda\delta)\|^2}
{4\lambda^2\|\delta\|^2},\qquad\lambda=1/64.
\label{eq:continuation-gain}
\end{equation}
Each of the four banks contains 1,024 new starts, four action branches, a separately sampled goal branch and H20 trajectories. Sixteen fixed checkpoints (three LeWM fine-tuning checkpoints and one PLDM checkpoint per task) see the same observations and actions. The LeWM checkpoints share a frozen encoder; both families use a ViT-tiny/14 encoder with 192 latent dimensions but different weights and predictors, so they are not independent replications of one backbone.

At H5, $P$ denotes the model's own prediction and $E$ the encoding of the realized image. In each of four substitution conditions $M$, the prediction's error is replaced by an empirical residual (a \emph{donor}) taken from a separate, earlier set of starts and rescaled to the same norm, so the current error size is unchanged and only its direction differs. The preceding predicted history, model weights and future actions are held fixed, and continuing from $P$ reproduces the model's original rollout. An independent CPU audit recomputes all inputs, norms, metrics and intervals. No zero residuals occurred.

The three primary outcomes are H6 MSE, H20 physical regret within the same four-candidate pool, and Equation~\ref{eq:continuation-gain}. We average over donors, branches and the LeWM checkpoints within each start before paired-start resampling, and adjust intervals over the 24 comparisons (Table~\ref{tab:resampling-settings}). Inference conditions on the fixed set of donors. The donor starts do not coincide with the new starts, which does not rule out overlap with pretraining episodes. The percentage reduction $(\mathrm{MSE}_P-\mathrm{MSE}_M)/\mathrm{MSE}_P$ uses aggregate means and is descriptive rather than a separately tested endpoint. The same banks are reused by the later probes and are not counted as additional independent starts.

\input{artifacts/continuation_confirmation_table_v1.tex}

\paragraph{Condition-matched directions.}
For each branch (the \emph{recipient}), the donors $L$ are the residuals of the other three action branches of the same start, and the donors $X$ are the residuals of the same branch indices at three other starts. Starts are randomly partitioned into 256 disjoint four-start groups, and the cross-start donors are cyclic permutations within each group, so the two donor collections have the same marginal distribution; every donor is rescaled to the recipient's own error norm. Four contrasts per model--task setting form a separate 32-comparison family, and resampling whole four-start groups preserves the dependence created by shared donors. Same-start donors are not matched physical states or measured tangent vectors.

\input{artifacts/continuation_conditional_gain_table_v1.tex}
\input{artifacts/continuation_conditional_mse_table_v1.tex}

\paragraph{Contraction and bias cancellation.}
With $y$ the true next encoding, $b=F(e)-y$ and $d_r=F(e+r)-F(e)$, next-step error is exactly
\begin{equation}
\|F(e+r)-y\|^2=\|b\|^2+\|d_r\|^2+2\langle b,d_r\rangle.
\label{eq:continuation-bias}
\end{equation}
LeWM/Reacher supplies a descriptive counterexample under true preceding
history: own-minus-same-start response energy is $+0.0006183$, while bias
alignment contributes $-0.0013301$, giving lower net MSE ($-0.0007118$).
Suppressing that response can therefore remove useful bias cancellation.

\paragraph{True-history control.}
Here the two preceding predicted latents $P_3,P_4$ are replaced by their true encodings $E_3,E_4$, while the current prediction $P_5$ (still generated from the predicted history), all substituted directions, scales, actions and weights are unchanged. The primary contrasts are $\kappa(L)-\kappa(X)$ under true history and its change relative to predicted history, forming a 16-comparison family with group resampling. All eight true-history contrasts remain positive; the changes comprise two positive, two negative and four unresolved intervals. Equation~\ref{eq:continuation-bias} is evaluated against the true next encoding $E_6$.

\input{artifacts/continuation_history_table_v1.tex}

\paragraph{Restricted Flow-JEPA extension~\citep{huo2026flowjepa}.}
Flow-JEPA generates a future segment jointly rather than one step at a time. On Reacher, we test two Flow-JEPA checkpoints with causal attention on 1,024 reused windows: at four intermediate steps the generated latent is replaced by $P$, $E$ or one of four equal-norm residuals from other windows, and generation continues, averaging over sixteen paired flow-noise draws per window. The sixteen next-step and final-output contrasts form one family. All eight next-step intervals favor the residuals from other windows; seven of eight final-output contrasts do so and one is unresolved (at the fourth step the two outputs coincide). This tests continuation from substituted intermediate latents on one task, not Flow-JEPA's native joint sampling or planning.

\paragraph{A fixed-reference continuation objective.}
We also asked whether training could exploit this sensitivity. Starting from three fixed 500-update base checkpoints per task, three objectives were each trained for another 500 updates (36 runs) with the same data, optimizer and update budget and with the encoder and projector frozen. The \emph{anchor} objective combines one-step MSE on expert and branch data with 0.5 times the H5 rollout MSE. The \emph{joint} objective adds 0.5 times the error of continuing each prediction $p_j$ with the true preceding history, measured against the true next encoding $e_{j+1}$. The \emph{fixed-reference} objective computes the same continuation error through a frozen copy of the initial network, so gradients reach the trained model only through its inputs; only the trained model is evaluated.

Evaluation uses H20 rollouts on the same four 1,024-start banks. The four prespecified endpoints (H6 MSE, H20 MSE, H20 response $Q$, H20 physical regret), two contrasts (fixed-reference minus anchor and minus joint) and four tasks form a 32-comparison paired-start family. Against joint, nine endpoints worsen and seven are unresolved; against anchor, three improve, one worsens and twelve are unresolved; all eight regret contrasts are unresolved. A stale reference network and useful joint adaptation are candidate explanations that these endpoints do not separate.

\paragraph{Closed-loop evaluation.}
A separately specified follow-up evaluates, per task, the original pretrained model and, for each of the three base initializations, the initial, anchor, joint and fixed-reference checkpoints (13 models) on 256 paired dataset episodes with native CEM (300 candidates, 30 iterations, top-30 elites, horizon H5 with action blocks of 5, goal offset 25 and an evaluation budget of 50). The two success-rate contrasts per task form an 8-comparison paired-episode family, averaging the three initializations within each episode first. Dataset episodes may overlap pretraining. Independent audits check every success against the planner's termination records, confirm that all models saw identical initial states, images and goals, and recompute all bootstrap replicates.

All eight primary CEM intervals include zero (Table~\ref{tab:fixed-cem-contrasts}), so the fixed-reference objective shows no confirmed benefit over anchor or joint training. Tables~\ref{tab:fixed-cem-rates}--\ref{tab:fixed-cem-counts} report the unmodified baseline and every initialization. No reference network or true future encoding is used at deployment.

\input{artifacts/fixed_continuation_cem_rates_v1.tex}
\input{artifacts/fixed_continuation_cem_contrasts_v1.tex}
\input{artifacts/fixed_continuation_cem_counts_v1.tex}

%% file: artifacts/continuation_confirmation_table_v1.tex
\begin{table}[!htbp]
\caption{All confirmation contrasts, $P-M$: fixed donors from an earlier bank, 1,024 starts/task, complete 24-comparison family. Positive MSE/regret means the own residual is worse; positive gain means greater squared response. Physical-score units differ across tasks. Bold marks adjusted intervals excluding zero.}
\label{tab:continuation-confirmation}
\centering\footnotesize
\setlength{\tabcolsep}{3pt}
\begin{tabular}{@{}lccc@{}}
\toprule
Model / task & $\Delta$ H6 MSE & $\Delta\kappa$ & $\Delta$ H20 regret \\
\midrule
LeWM / PushT & \textbf{0.09760} \ci{0.08369}{0.11317} & \textbf{0.15049} \ci{0.13037}{0.17288} & \textbf{0.40443} \ci{0.24743}{0.57578} \\
LeWM / TwoRoom & \textbf{0.01509} \ci{0.01105}{0.02126} & \textbf{0.47289} \ci{0.42852}{0.51998} & 0.00686 \ci{-0.00728}{0.02228} \\
LeWM / Reacher & \textbf{0.00239} \ci{0.00142}{0.00339} & \textbf{0.11995} \ci{0.10766}{0.13297} & 0.05231 \ci{-0.14183}{0.25165} \\
LeWM / Cube & \textbf{0.04657} \ci{0.03544}{0.05851} & \textbf{0.25722} \ci{0.22777}{0.28822} & -0.00132 \ci{-0.02446}{0.02031} \\
PLDM / PushT & \textbf{0.17455} \ci{0.14686}{0.20350} & \textbf{0.32276} \ci{0.27144}{0.38276} & \textbf{0.18968} \ci{0.00861}{0.38629} \\
PLDM / TwoRoom & \textbf{0.03597} \ci{0.02470}{0.05042} & \textbf{0.46066} \ci{0.40792}{0.52052} & 0.01753 \ci{-0.00385}{0.04778} \\
PLDM / Reacher & \textbf{0.37402} \ci{0.31817}{0.42927} & \textbf{0.95594} \ci{0.83456}{1.08896} & 0.09870 \ci{-0.24658}{0.44611} \\
PLDM / Cube & \textbf{0.34831} \ci{0.30516}{0.39376} & \textbf{0.53807} \ci{0.47920}{0.59982} & 0.01219 \ci{-0.03815}{0.06546} \\
\bottomrule
\end{tabular}
\end{table}

%% file: artifacts/continuation_conditional_gain_table_v1.tex
\begin{table}[!htbp]
\caption{Condition-matched squared directional gains (32-comparison family; 256 four-start groups). Same-start $L$ exceeds cross-start $X$ in all eight settings; an unresolved $P-L$ contrast is not equality.}
\label{tab:continuation-conditional_gain}
\centering\footnotesize
\setlength{\tabcolsep}{3pt}
\begin{tabular}{@{}lcc@{}}
\toprule
Model / task & $\kappa(P)-\kappa(L)$ & $\kappa(L)-\kappa(X)$ \\
\midrule
LeWM / PushT & 0.00637 \ci{-0.00868}{0.02078} & \textbf{0.13593} \ci{0.11516}{0.15885} \\
LeWM / TwoRoom & \textbf{0.09405} \ci{0.07493}{0.11516} & \textbf{0.37594} \ci{0.33789}{0.41538} \\
LeWM / Reacher & \textbf{0.01826} \ci{0.01558}{0.02109} & \textbf{0.10342} \ci{0.09365}{0.11354} \\
LeWM / Cube & \textbf{0.04302} \ci{0.03167}{0.05474} & \textbf{0.22768} \ci{0.19922}{0.25816} \\
PLDM / PushT & \textbf{0.04572} \ci{0.01504}{0.07566} & \textbf{0.26685} \ci{0.21957}{0.32254} \\
PLDM / TwoRoom & \textbf{0.08714} \ci{0.06705}{0.11183} & \textbf{0.37484} \ci{0.33200}{0.42442} \\
PLDM / Reacher & \textbf{0.24693} \ci{0.18588}{0.31459} & \textbf{0.68497} \ci{0.59181}{0.78844} \\
PLDM / Cube & \textbf{0.09230} \ci{0.06714}{0.11919} & \textbf{0.44650} \ci{0.39847}{0.50134} \\
\bottomrule
\end{tabular}
\end{table}

%% file: artifacts/continuation_conditional_mse_table_v1.tex
\begin{table}[!htbp]
\caption{Condition-matched H6 MSE differences, retaining the same 32-comparison family. The LeWM/Reacher $P-L$ reversal is reported, not removed.}
\label{tab:continuation-conditional_mse}
\centering\footnotesize
\setlength{\tabcolsep}{3pt}
\begin{tabular}{@{}lcc@{}}
\toprule
Model / task & MSE$(P)-$MSE$(L)$ & MSE$(L)-$MSE$(X)$ \\
\midrule
LeWM / PushT & \textbf{0.03390} \ci{0.02521}{0.04346} & \textbf{0.06377} \ci{0.05404}{0.07492} \\
LeWM / TwoRoom & \textbf{0.00514} \ci{0.00261}{0.00902} & \textbf{0.01007} \ci{0.00781}{0.01336} \\
LeWM / Reacher & \textbf{-0.00053} \ci{-0.00083}{-0.00024} & \textbf{0.00278} \ci{0.00189}{0.00370} \\
LeWM / Cube & \textbf{0.00690} \ci{0.00274}{0.01162} & \textbf{0.04122} \ci{0.03063}{0.05193} \\
PLDM / PushT & \textbf{0.05397} \ci{0.03888}{0.06953} & \textbf{0.11496} \ci{0.09392}{0.13698} \\
PLDM / TwoRoom & \textbf{0.00774} \ci{0.00418}{0.01241} & \textbf{0.02878} \ci{0.01945}{0.04063} \\
PLDM / Reacher & \textbf{0.09164} \ci{0.07444}{0.10982} & \textbf{0.28328} \ci{0.23459}{0.33122} \\
PLDM / Cube & \textbf{0.07907} \ci{0.06332}{0.09524} & \textbf{0.27150} \ci{0.23445}{0.30992} \\
\bottomrule
\end{tabular}
\end{table}

%% file: artifacts/continuation_history_table_v1.tex
\begin{table}[!htbp]
\caption{True-history control (16-comparison family; 256 four-start groups). All $L-X$ gains remain positive, but changes relative to predicted history do not establish equivalence.}
\label{tab:continuation-history}
\centering\footnotesize
\setlength{\tabcolsep}{3pt}
\begin{tabular}{@{}lcc@{}}
\toprule
Model / task & True-history $\kappa(L)-\kappa(X)$ & True minus predicted history \\
\midrule
LeWM / PushT & \textbf{0.13669} \ci{0.11662}{0.15862} & 0.00076 \ci{-0.00058}{0.00230} \\
LeWM / TwoRoom & \textbf{0.37571} \ci{0.34073}{0.41245} & \textbf{-0.00023} \ci{-0.00048}{-0.00000} \\
LeWM / Reacher & \textbf{0.10458} \ci{0.09516}{0.11400} & \textbf{0.00117} \ci{0.00033}{0.00200} \\
LeWM / Cube & \textbf{0.23265} \ci{0.20535}{0.26142} & \textbf{0.00498} \ci{0.00214}{0.00802} \\
PLDM / PushT & \textbf{0.25863} \ci{0.21590}{0.30631} & -0.00821 \ci{-0.01854}{0.00037} \\
PLDM / TwoRoom & \textbf{0.37458} \ci{0.33537}{0.41937} & -0.00026 \ci{-0.00086}{0.00035} \\
PLDM / Reacher & \textbf{0.67881} \ci{0.59155}{0.77119} & -0.00617 \ci{-0.05957}{0.04819} \\
PLDM / Cube & \textbf{0.43131} \ci{0.38645}{0.47894} & \textbf{-0.01519} \ci{-0.03054}{-0.00110} \\
\bottomrule
\end{tabular}
\end{table}

%% file: artifacts/fixed_continuation_cem_rates_v1.tex
\begin{table}[t]
\centering\small
\caption{Native-CEM success (\%). Original is the unmodified pretrained checkpoint; other columns average three fixed fine-tuning initializations on the same 256 episodes per task. Fixed denotes fixed-reference training, with only the student deployed. These rates are descriptive, not separate significance tests.}
\label{tab:fixed-cem-rates}
\begin{tabular}{lrrrrr}
\toprule
Task & Original & Initial & Anchor & Joint & Fixed \\
\midrule
PushT & 90.62 & 84.90 & 84.64 & 83.46 & 84.77 \\
TwoRoom & 85.16 & 85.16 & 84.51 & 84.64 & 84.90 \\
Reacher & 58.59 & 56.90 & 62.11 & 59.24 & 59.90 \\
Cube & 68.36 & 70.44 & 73.57 & 72.27 & 72.53 \\
\bottomrule
\end{tabular}
\end{table}

%% file: artifacts/fixed_continuation_cem_contrasts_v1.tex
\begin{table}[t]
\centering\small
\caption{All eight predeclared fixed-reference CEM success contrasts, in percentage points (positive favors Fixed). Intervals use 50,000 paired-episode bootstrap draws and a Bonferroni correction over the eight comparisons (99.375\% per interval). All include zero: no confirmed benefit or equivalence.}
\label{tab:fixed-cem-contrasts}
\begin{tabular}{llrrl}
\toprule
Task & Contrast & Difference & Adjusted interval & Status \\
\midrule
PushT & Fixed-anchor & +0.13 & $[-2.47, +2.73]$ & Unresolved \\
PushT & Fixed-joint & +1.30 & $[-1.43, +4.17]$ & Unresolved \\
TwoRoom & Fixed-anchor & +0.39 & $[-1.04, +1.82]$ & Unresolved \\
TwoRoom & Fixed-joint & +0.26 & $[-0.91, +1.56]$ & Unresolved \\
Reacher & Fixed-anchor & -2.21 & $[-6.90, +2.60]$ & Unresolved \\
Reacher & Fixed-joint & +0.65 & $[-4.30, +5.47]$ & Unresolved \\
Cube & Fixed-anchor & -1.04 & $[-3.39, +1.17]$ & Unresolved \\
Cube & Fixed-joint & +0.26 & $[-2.34, +2.73]$ & Unresolved \\
\bottomrule
\end{tabular}
\end{table}

%% file: artifacts/fixed_continuation_cem_counts_v1.tex
\begin{table}[t]
\centering\small
\caption{All fine-tuning-initialization CEM success counts, each out of the same 256 episodes per task. Original counts are PushT 232, TwoRoom 218, Reacher 150 and Cube 175. Repeated models do not add independent physical episodes.}
\label{tab:fixed-cem-counts}
\begin{tabular}{lrrrrr}
\toprule
Task & Initialization & Initial & Anchor & Joint & Fixed \\
\midrule
PushT & 94111 & 213 & 220 & 215 & 214 \\
PushT & 94112 & 213 & 215 & 218 & 220 \\
PushT & 94113 & 226 & 215 & 208 & 217 \\
TwoRoom & 94111 & 219 & 216 & 218 & 216 \\
TwoRoom & 94112 & 216 & 218 & 216 & 218 \\
TwoRoom & 94113 & 219 & 215 & 216 & 218 \\
Reacher & 94111 & 144 & 150 & 157 & 148 \\
Reacher & 94112 & 152 & 159 & 145 & 160 \\
Reacher & 94113 & 141 & 168 & 153 & 152 \\
Cube & 94111 & 182 & 186 & 183 & 184 \\
Cube & 94112 & 177 & 191 & 186 & 186 \\
Cube & 94113 & 182 & 188 & 186 & 187 \\
\bottomrule
\end{tabular}
\end{table}

%% file: continuation_secant_appendix.tex
\subsection{Action-secant geometry and component interventions}
\label{app:secant}

The direction effect above does not identify its training origin or a uniquely harmful latent component. We therefore report the full sequence of exploratory geometry and intervention tests, including predictions that were not supported. All use the same four banks: 1,024 starts per task, four branches, LeWM/PLDM and the sixteen fixed checkpoints, i.e., 4,096 physical starts shared across model families rather than 8,192 new starts. Each protocol that required new model evaluations was fixed before running them, but later protocols were designed after seeing earlier results. The three comparison families below are separate, sequential explorations rather than one independently confirmed mechanism. All average fixed checkpoints and branches before resampling the same 256 four-start groups (Table~\ref{tab:resampling-settings}).

\subsubsection{Projection onto held-out action secants}

For recipient branch $j$ at H5, form the span of differences among the other three realized endpoint encodings. Its orthogonal projector $\Pi_j$ has rank at most two; excluding the recipient avoids directly sharing the $-e_j$ term with $\delta_j=p_j-e_j$. Define $f_j=\|\Pi_j\delta_j\|^2/\|\delta_j\|^2$. This is a finite action-sequence secant span, potentially including observation geometry, not a recovered physical tangent space. The cross-start control uses the other starts in the matched four-start groups, preserving the empirical direction marginal. In an 8-comparison family, the own-minus-cross-start projection fraction is positive in all eight settings, by 11.79--31.44 percentage points (Table~\ref{tab:secant-geometry}). Rank-zero and rank-one cases are retained: per checkpoint, PushT has 216/252/3,628 recipient branches of ranks 0/1/2, TwoRoom 71/45/3,980, and Reacher/Cube 0/0/4,096. No zero total residual occurs. Geometry alone does not establish harmful propagation or mediation.

\input{artifacts/secant_geometry_table_v2.tex}

\subsubsection{Matched component interventions}

Write $\delta_T=\Pi_j\delta$ and $\delta_N=\delta-\delta_T$. With the true preceding history $E_3,E_4$ and the next action fixed, compare $N=e+\delta_N$ to $A$, the full direction $\delta$ rescaled to the same remaining input-error norm; compare $T=e+\delta_T$ to its own norm-matched full-direction control $B$. The original prediction $P$ and true encoding $E$ are anchors. Comparisons are within $N/A$ and $T/B$, not between $N$ and $T$, whose error budgets differ. The current $P_5$ still originates from predicted history. Response energy is $\|F(z)-F(e)\|^2/192$, and MSE is measured against the true encoding $E_6$.

Four contrasts per setting form a 32-comparison family. Removing the parallel component reduces response energy in all eight $N-A$ comparisons and MSE in six (two unresolved). The reciprocal prediction, that retaining only this component should increase energy relative to $B$, is not supported: four contrasts are negative and four unresolved (Tables~\ref{tab:secant-energy}--\ref{tab:secant-mse}). The favorable removal result therefore does not identify a separately harmful component. PLDM/Reacher also shows that contraction need not reduce error: $T-B$ energy is $-0.092806$, but MSE is $+0.036054$, with both adjusted intervals excluding zero; the bias cross-term difference is $+0.128860$ (descriptive algebraic decomposition).

\input{artifacts/secant_energy_table_v2.tex}
\input{artifacts/secant_mse_table_v2.tex}

\paragraph{Post-hoc finite-response identity.}
After inspecting these 32 outcomes, we decomposed $r_m=F(z_m)-F(e)$ and $\eta=r_P-r_N-r_T$ without adding hypothesis tests:
\begin{equation}
\frac{\|r_P\|^2-\|r_N\|^2-\|r_T\|^2}{192}
=\frac{2\langle r_N,r_T\rangle}{192}
+\frac{2\langle r_N+r_T,\eta\rangle+\|\eta\|^2}{192}.
\end{equation}
Input cross-term means are near zero, but output cross-term means are positive in all eight settings. The finite nonadditive correction is negative in all eight and reverses the total excess in two (Table~\ref{tab:secant-posthoc}). This observation alone establishes neither a local-Jacobian mechanism nor action specificity; it motivated the following control.

\input{artifacts/secant_posthoc_table_v2.tex}

\clearpage
\subsubsection{Small-scale, energy-matched random decompositions}

For $R=\|\delta\|$, $f=\|\delta_T\|^2/R^2$ and random unit $v\perp\delta$, set
\begin{equation}
T_R=f\delta+\sqrt{f(1-f)}Rv,\qquad N_R=\delta-T_R.
\end{equation}
This preserves the total residual, orthogonality and both component energies while changing the decomposition's direction. Three fixed Gaussian draws are orthogonalized to $\delta$ and averaged, not selected. Degenerate $f=0/1$ is retained; no numerical boundary clipping or zero residual occurred. Directions enter the same fixed continuation with true preceding history. For $D_v=[F(e+\lambda v)-F(e-\lambda v)]/(2\lambda)$, the primary quantity is $2\langle D_T,D_N\rangle/\|\delta\|^2$, normalized per latent before averaging. The two contrasts per setting are real coupling and real minus mean random coupling, with $\lambda=1/64$ and a 16-comparison family.

Real coupling is positive in seven settings and unresolved in one, but real-minus-random coupling is negative in all eight (Table~\ref{tab:secant-coupling}). The test therefore does not support an action-specific excess-coupling explanation. Whether random finite-amplitude correction would improve MSE or CEM was not tested. At the diagnostic scale $\lambda=1/128$, no contrast mean changes sign, and the relative L2 difference over per-latent coupling values is 0.0198--0.0853\%. These are finite-difference scale checks rather than exact-Jacobian or per-latent sign-invariance claims.

\input{artifacts/secant_coupling_table_v2.tex}

\paragraph{Summary.}
These tests retain the condition-associated direction effect and the measured $N/A$ differences, while not supporting two stronger component explanations. They do not establish disjoint P/E domains, a full tangent space or why training creates the association, and the fixed-reference objective's eight native-CEM comparisons remain unresolved. No secant result is attributed to Flow-JEPA. Independent CPU audits reconstruct all 56 reported contrasts.

%% file: artifacts/secant_geometry_table_v2.tex
\begin{table}[!htbp]
\caption{Held-out action-secant projection fractions. Own residual $P$ versus cross-start residual $X$, with the recipient branch excluded from the reference span. All eight $P-X$ contrasts have positive intervals (8-comparison family). Fractions, not percentages; this finite span is not a full physical tangent space.}
\label{tab:secant-geometry}
\centering\footnotesize
\setlength{\tabcolsep}{3pt}
\begin{tabular}{@{}lccc@{}}
\toprule
Model / task & $f_P$ & $f_X$ & $f_P-f_X$ \\
\midrule
LeWM / PushT & 0.2419 & 0.0769 & \textbf{0.164945} \ci{0.152492}{0.177439} \\
LeWM / TwoRoom & 0.3646 & 0.0527 & \textbf{0.311914} \ci{0.298531}{0.325415} \\
LeWM / Reacher & 0.2467 & 0.1287 & \textbf{0.117934} \ci{0.107290}{0.128552} \\
LeWM / Cube & 0.2367 & 0.0395 & \textbf{0.197243} \ci{0.186697}{0.208054} \\
PLDM / PushT & 0.2214 & 0.0592 & \textbf{0.162224} \ci{0.151556}{0.173043} \\
PLDM / TwoRoom & 0.3445 & 0.0301 & \textbf{0.314367} \ci{0.297506}{0.331499} \\
PLDM / Reacher & 0.2519 & 0.0176 & \textbf{0.234261} \ci{0.221346}{0.247353} \\
PLDM / Cube & 0.3071 & 0.0176 & \textbf{0.289455} \ci{0.276927}{0.302020} \\
\bottomrule
\end{tabular}
\end{table}

%% file: artifacts/secant_energy_table_v2.tex
\begin{table}[!htbp]
\caption{Finite component interventions, response energy: all 16 contrasts of the 32-comparison family. Negative $N-A$ was expected; positive $T-B$ was expected but is supported in no setting. Bold denotes an adjusted interval excluding zero, irrespective of the expected sign.}
\label{tab:secant-energy}
\centering\footnotesize
\setlength{\tabcolsep}{3pt}
\begin{tabular}{@{}lcc@{}}
\toprule
Model / task & $\Delta$ energy $N-A$ & $\Delta$ energy $T-B$ \\
\midrule
LeWM / PushT & \textbf{-0.007923} \ci{-0.010538}{-0.005528} & \textbf{-0.003829} \ci{-0.005957}{-0.001738} \\
LeWM / TwoRoom & \textbf{-0.002766} \ci{-0.005804}{-0.001479} & -0.000556 \ci{-0.001896}{0.000238} \\
LeWM / Reacher & \textbf{-0.000820} \ci{-0.001018}{-0.000630} & 0.000136 \ci{-0.000035}{0.000300} \\
LeWM / Cube & \textbf{-0.007049} \ci{-0.009365}{-0.004987} & -0.001851 \ci{-0.004239}{0.000105} \\
PLDM / PushT & \textbf{-0.032906} \ci{-0.041168}{-0.025480} & \textbf{-0.011992} \ci{-0.017771}{-0.006405} \\
PLDM / TwoRoom & \textbf{-0.005907} \ci{-0.009035}{-0.003747} & -0.001383 \ci{-0.003707}{0.000268} \\
PLDM / Reacher & \textbf{-0.084467} \ci{-0.096655}{-0.073149} & \textbf{-0.092806} \ci{-0.114898}{-0.072683} \\
PLDM / Cube & \textbf{-0.012675} \ci{-0.021800}{-0.003650} & \textbf{-0.058227} \ci{-0.072235}{-0.045113} \\
\bottomrule
\end{tabular}
\end{table}

%% file: artifacts/secant_mse_table_v2.tex
\begin{table}[!htbp]
\caption{Finite component interventions, next-step MSE: the other 16 contrasts of the same 32-comparison family. PLDM/Reacher $T-B$ has lower response energy but higher MSE; stronger contraction is not automatically more accurate. Raw latent units are not comparable across encoders.}
\label{tab:secant-mse}
\centering\footnotesize
\setlength{\tabcolsep}{3pt}
\begin{tabular}{@{}lcc@{}}
\toprule
Model / task & $\Delta$ MSE $N-A$ & $\Delta$ MSE $T-B$ \\
\midrule
LeWM / PushT & \textbf{-0.007899} \ci{-0.010986}{-0.004895} & \textbf{-0.007218} \ci{-0.010145}{-0.004299} \\
LeWM / TwoRoom & \textbf{-0.002800} \ci{-0.005834}{-0.001485} & -0.000578 \ci{-0.001976}{0.000240} \\
LeWM / Reacher & 0.000076 \ci{-0.000294}{0.000449} & 0.000020 \ci{-0.000353}{0.000387} \\
LeWM / Cube & \textbf{-0.004531} \ci{-0.007945}{-0.001152} & 0.002280 \ci{-0.001649}{0.006165} \\
PLDM / PushT & \textbf{-0.015862} \ci{-0.025823}{-0.006149} & \textbf{-0.008323} \ci{-0.016075}{-0.000516} \\
PLDM / TwoRoom & \textbf{-0.006255} \ci{-0.009440}{-0.003940} & -0.000613 \ci{-0.002971}{0.001222} \\
PLDM / Reacher & \textbf{-0.054640} \ci{-0.067216}{-0.042779} & \textbf{0.036054} \ci{0.012216}{0.060075} \\
PLDM / Cube & 0.000404 \ci{-0.010041}{0.010814} & \textbf{-0.063805} \ci{-0.082606}{-0.044662} \\
\bottomrule
\end{tabular}
\end{table}

%% file: artifacts/secant_posthoc_table_v2.tex
\begin{table}[!htbp]
\caption{Post-hoc descriptive finite-response decomposition. Cross terms are positive in all eight sample means, but nonlinear corrections are negative throughout and total excess is negative for PLDM/Reacher and Cube. No confidence intervals or additional hypothesis tests were introduced.}
\label{tab:secant-posthoc}
\centering\footnotesize
\setlength{\tabcolsep}{3pt}
\begin{tabular}{@{}lccc@{}}
\toprule
Model / task & Response cross term & Nonadditive correction & Total excess \\
\midrule
LeWM / PushT & 0.016418 & -0.010174 & 0.006244 \\
LeWM / TwoRoom & 0.002928 & -0.000907 & 0.002020 \\
LeWM / Reacher & 0.000711 & -0.000396 & 0.000315 \\
LeWM / Cube & 0.007868 & -0.002043 & 0.005826 \\
PLDM / PushT & 0.048264 & -0.030126 & 0.018138 \\
PLDM / TwoRoom & 0.006832 & -0.002144 & 0.004688 \\
PLDM / Reacher & 0.223139 & -0.234933 & -0.011794 \\
PLDM / Cube & 0.112563 & -0.136532 & -0.023969 \\
\bottomrule
\end{tabular}
\end{table}

%% file: artifacts/secant_coupling_table_v2.tex
\begin{table}[!htbp]
\caption{Small-scale coupling versus energy-matched random orthogonal decompositions ($\lambda=1/64$; 16-comparison family). Real action-secant coupling is smaller in all eight settings. Positive real coupling alone does not establish action-specific excess coupling.}
\label{tab:secant-coupling}
\centering\footnotesize
\setlength{\tabcolsep}{3pt}
\begin{tabular}{@{}lcc@{}}
\toprule
Model / task & Real coupling & Real minus random \\
\midrule
LeWM / PushT & 0.004066 \ci{-0.003901}{0.011637} & \textbf{-0.140713} \ci{-0.150916}{-0.131311} \\
LeWM / TwoRoom & \textbf{0.082070} \ci{0.064108}{0.099741} & \textbf{-0.127948} \ci{-0.144895}{-0.112618} \\
LeWM / Reacher & \textbf{0.017043} \ci{0.011433}{0.022641} & \textbf{-0.072477} \ci{-0.076673}{-0.068296} \\
LeWM / Cube & \textbf{0.031253} \ci{0.020827}{0.041818} & \textbf{-0.086397} \ci{-0.095481}{-0.077789} \\
PLDM / PushT & \textbf{0.044083} \ci{0.024719}{0.065862} & \textbf{-0.087057} \ci{-0.101807}{-0.073359} \\
PLDM / TwoRoom & \textbf{0.071182} \ci{0.055293}{0.087785} & \textbf{-0.096474} \ci{-0.111827}{-0.082444} \\
PLDM / Reacher & \textbf{0.170522} \ci{0.128087}{0.216095} & \textbf{-0.106169} \ci{-0.130705}{-0.082598} \\
PLDM / Cube & \textbf{0.054756} \ci{0.032129}{0.077155} & \textbf{-0.120280} \ci{-0.144393}{-0.100520} \\
\bottomrule
\end{tabular}
\end{table}

%% file: residual_origin_appendix.tex
\subsection{Residual origin and specificity controls}
\label{app:direction-specificity}

\paragraph{Question and chronology.}
These follow-ups reuse the four 1,024-start, four-branch banks and
sixteen fixed checkpoints of Appendix~\ref{app:continuation}; they are
exploratory rather than an additional independent confirmation. Both
families share each physical bank, and the three LeWM fine-tuning
checkpoints share an encoder and are averaged before resampling. They ask
whether the positive directional signature is specific to predictor error.

In a locally linear reference with a constant map $J$, permuting a marginally
matched collection of normalized directions leaves average
$\|J\delta\|^2/\|\delta\|^2$ unchanged. A positive $L-X$ gap therefore indicates
an association with the conditional map beyond this constant-anisotropy case,
not that the association is exclusive to errors produced by a predictor.

\paragraph{Where the residual is generated.}
The teacher-origin test compares the model's own residual $P_5-E_5$ with the
residual of a single prediction made from the true causal history. Both
supply directions rescaled to the recipient's own error norm. During the probe, the preceding
$E_3,E_4$, current center $E_5$, next action and model parameters are fixed.
The seven directions are own, three same-start other-branch donors $L$, and
three cross-start donors $X$. Donors are permuted within 256 disjoint groups
of four starts, preserving the marginal donor collection; the groups, not
branches or checkpoint replicas, are bootstrap units. No zero residual
directions occurred. At $\lambda=1/64$ these contrasts form a 16-comparison
family (Table~\ref{tab:resampling-settings}); half scale is a diagnostic only.

Table~\ref{tab:priority-teacher} reports every primary contrast. All eight
teacher $L-X$ gaps are positive; teacher minus native is negative in six and
unresolved in two. The direction association thus arises without a
self-generated input history and is strengthened by that history in six
tested settings. The point ratio of teacher/native gaps is not a mediated
causal fraction.

\input{artifacts/priority_teacher_table_v2.tex}

\paragraph{Controls that do not require a learned predictor residual.}
Persistence uses $E_0-E_5$, the reverse of a real cumulative trajectory change.
Centroid uses $\mu_{\mathrm{old}}-E_5$, where $\mu_{\mathrm{old}}$ is computed
from the true H5 encodings of a separate, earlier bank and held fixed for
all current groups. It is not estimated from the recipient group or its peers.
Each control supplies own/$L$/$X$ directions normalized to the recipient's own error
norm; continuation history, action, center and model remain unchanged.

Zero directions are excluded rather than replaced by an artificial direction.
PushT has 19 zero persistence directions in five groups, leaving 251 groups
and 1,004 starts; teacher comparisons use exactly those same groups. All other
persistence and all centroid cases use 256 groups. At $\lambda=1/64$ the
persistence and centroid contrasts form a 32-comparison family; each control
uses the same bootstrap seed on its eligible groups, and the fixed external
centroid is conditioned upon rather than re-estimated in each draw.

Tables~\ref{tab:priority-persistence}--\ref{tab:priority-centroid} report all 32
contrasts. Persistence $L-X$ is positive in all eight settings, and teacher
minus persistence is three positive, four negative and one unresolved.
Centroid $L-X$ has six positive, one negative and one unresolved contrast;
teacher minus centroid is four positive, one negative and three unresolved.
Half-scale diagnostic point estimates have no sign flips. Real-trajectory
directions therefore reproduce the positive-gain signature, so it is not
specific to predictor error; the inserted directions need not be physically
compatible with the fixed preceding history.

\input{artifacts/priority_persistence_table_v2.tex}
\input{artifacts/priority_centroid_table_v2.tex}

%% file: artifacts/priority_teacher_table_v2.tex
\begin{table}[!htbp]
\caption{True-history residual generation: all 16 contrasts (16-comparison family). The continuation map and each recipient error norm remain fixed. Bold indicates an adjusted interval excluding zero, not necessarily the hypothesized sign.}
\label{tab:priority-teacher}
\centering\footnotesize
\setlength{\tabcolsep}{3pt}
\begin{tabular}{@{}lcc@{}}
\toprule
Model / task & Teacher $L-X$ & Teacher minus native $L-X$ \\
\midrule
LeWM / PushT & \textbf{0.112495} \ci{0.085692}{0.143114} & -0.024197 \ci{-0.049751}{0.005279} \\
LeWM / TwoRoom & \textbf{0.214829} \ci{0.184216}{0.247567} & \textbf{-0.160880} \ci{-0.188319}{-0.133072} \\
LeWM / Reacher & \textbf{0.057607} \ci{0.051156}{0.064272} & \textbf{-0.046976} \ci{-0.055577}{-0.038620} \\
LeWM / Cube & \textbf{0.194810} \ci{0.167430}{0.223418} & \textbf{-0.037843} \ci{-0.060604}{-0.013599} \\
PLDM / PushT & \textbf{0.210159} \ci{0.173971}{0.252970} & \textbf{-0.048474} \ci{-0.090733}{-0.008256} \\
PLDM / TwoRoom & \textbf{0.203771} \ci{0.174429}{0.236751} & \textbf{-0.170810} \ci{-0.204543}{-0.141825} \\
PLDM / Reacher & \textbf{0.623899} \ci{0.534634}{0.722817} & -0.054907 \ci{-0.138244}{0.030548} \\
PLDM / Cube & \textbf{0.365088} \ci{0.324359}{0.409832} & \textbf{-0.066221} \ci{-0.100659}{-0.030215} \\
\bottomrule
\end{tabular}
\end{table}

%% file: artifacts/priority_persistence_table_v2.tex
\begin{table}[!htbp]
\caption{Negative control from real cumulative change, all 16 associated contrasts of the 32-comparison family. Persistence directions are $E_0-E_5$, not learned predictor residuals. PushT uses 251 valid four-start groups for both arms; other tasks use 256.}
\label{tab:priority-persistence}
\centering\footnotesize
\setlength{\tabcolsep}{3pt}
\begin{tabular}{@{}lcc@{}}
\toprule
Model / task & Persistence $L-X$ & Teacher minus persistence \\
\midrule
LeWM / PushT & \textbf{0.067476} \ci{0.040936}{0.095913} & \textbf{0.046232} \ci{0.012818}{0.083248} \\
LeWM / TwoRoom & \textbf{0.334933} \ci{0.274951}{0.398615} & \textbf{-0.120104} \ci{-0.182330}{-0.060132} \\
LeWM / Reacher & \textbf{0.176905} \ci{0.164643}{0.189121} & \textbf{-0.119298} \ci{-0.131259}{-0.107234} \\
LeWM / Cube & \textbf{0.188272} \ci{0.149008}{0.228418} & 0.006538 \ci{-0.036826}{0.050508} \\
PLDM / PushT & \textbf{0.305623} \ci{0.245357}{0.376711} & \textbf{-0.095089} \ci{-0.163596}{-0.033823} \\
PLDM / TwoRoom & \textbf{0.409728} \ci{0.346408}{0.477856} & \textbf{-0.205957} \ci{-0.270396}{-0.143363} \\
PLDM / Reacher & \textbf{0.379127} \ci{0.326259}{0.433949} & \textbf{0.244771} \ci{0.142218}{0.358036} \\
PLDM / Cube & \textbf{0.281112} \ci{0.231907}{0.338633} & \textbf{0.083976} \ci{0.029083}{0.140331} \\
\bottomrule
\end{tabular}
\end{table}

%% file: artifacts/priority_centroid_table_v2.tex
\begin{table}[!htbp]
\caption{Fixed center of an independent bank minus $E_5$: the remaining 16 contrasts of the 32-comparison family. All cases use 256 groups. Positive conditional directional gain is not a predictor-error-specific signature.}
\label{tab:priority-centroid}
\centering\footnotesize
\setlength{\tabcolsep}{3pt}
\begin{tabular}{@{}lcc@{}}
\toprule
Model / task & Centroid $L-X$ & Teacher minus centroid \\
\midrule
LeWM / PushT & \textbf{-0.087113} \ci{-0.103184}{-0.070151} & \textbf{0.199608} \ci{0.168100}{0.234424} \\
LeWM / TwoRoom & \textbf{0.078492} \ci{0.019825}{0.141365} & \textbf{0.136337} \ci{0.074545}{0.196322} \\
LeWM / Reacher & \textbf{0.351561} \ci{0.333254}{0.369986} & \textbf{-0.293955} \ci{-0.310126}{-0.277841} \\
LeWM / Cube & \textbf{0.105642} \ci{0.077123}{0.135579} & \textbf{0.089168} \ci{0.051830}{0.127332} \\
PLDM / PushT & -0.007828 \ci{-0.041842}{0.026288} & \textbf{0.217987} \ci{0.167479}{0.273493} \\
PLDM / TwoRoom & \textbf{0.221413} \ci{0.169795}{0.273275} & -0.017642 \ci{-0.070652}{0.036955} \\
PLDM / Reacher & \textbf{0.657461} \ci{0.607060}{0.708809} & -0.033563 \ci{-0.130471}{0.069393} \\
PLDM / Cube & \textbf{0.404397} \ci{0.348253}{0.464615} & -0.039309 \ci{-0.090293}{0.012730} \\
\bottomrule
\end{tabular}
\end{table}

%% file: collapse_appendix.tex
\section{Representation collapse as a positive control}
\label{app:collapse}

\begin{figure}[!htbp]
    \centering
    \includegraphics[width=\linewidth]{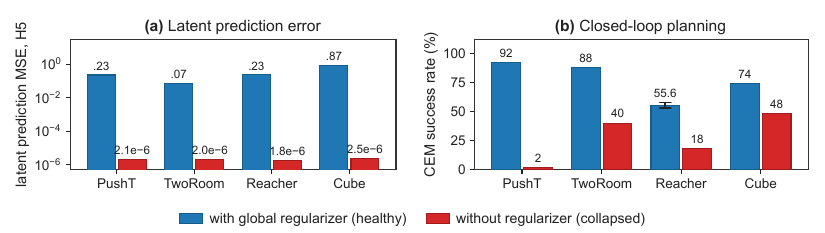}
    \caption{\textbf{Global representation collapse makes prediction look easy while planning degrades.} Removing the global regularizer lowers H5 latent prediction MSE by more than four orders of magnitude (a) while closed-loop CEM success falls on every task (b). Historical evaluation sizes are heterogeneous (mostly one 50-episode seed; healthy Reacher five seeds, bar $\pm$1 s.e.): a motivating positive control, not a paired confirmatory experiment.}
    \label{fig:collapse}
\end{figure}

Removing LeWM's Gaussian regularizer supplies an extreme positive control (Figure~\ref{fig:collapse}). H5 raw latent prediction MSE falls to $1.8\times10^{-6}$--$2.5\times10^{-6}$ across the four tasks, while closed-loop CEM success is 2\%, 40\%, 18\%, and 48\%, compared with 92\%, 88\%, 55.6\%, and 74\% for the corresponding healthy models. This historical comparison illustrates how target-space collapse can make latent prediction vacuous. Its heterogeneous evaluation sizes make it descriptive: most checkpoints use one 50-episode seed, while healthy Reacher uses a later five-seed re-evaluation. The main text concerns healthy checkpoints, whose prediction error is informative in aggregate but not about action selection.

